\documentclass[sigconf]{acmart}

\setcopyright{none}
\renewcommand\footnotetextcopyrightpermission[1]{} 
\AtBeginDocument{%
  }
    
\usepackage{graphicx}
\usepackage{colortbl}
\usepackage{array}
\usepackage{multirow}
\usepackage{hhline}

\usepackage{longtable} 
\usepackage{array} 
\usepackage{booktabs} 
\usepackage{ragged2e} 
\newcolumntype{L}[1]{>{\RaggedRight\arraybackslash\hspace{0pt}}p{#1}} 

\definecolor{Gray}{gray}{0.85}
\definecolor{LightBlue}{rgb}{0.8,0.9,1} 
\definecolor{LightRed}{rgb}{1,0.8,0.8} 

\usepackage[most]{tcolorbox}

\definecolor{highlight}{RGB}{255,255,0} 
\definecolor{highlightblue}{RGB}{204, 229, 255}  
\definecolor{highlightorange}{RGB}{255, 229, 204} 
\definecolor{highlightred}{RGB}{255, 204, 204}      
\definecolor{highlightgreen}{RGB}{204, 255, 204}  
\definecolor{highlightLightPurple}{RGB}{229,204,255}

\setcopyright{acmlicensed}
\copyrightyear{2024}
\acmYear{2024}
\acmDOI{XXXXXXX.XXXXXXX}

\acmConference[30th]{ACM SIGKDD Conference on Knowledge Discovery and Data Mining}{ACM KDD August 25 - 29, 2024,
  2024}{Barcelona, Spain}
\acmISBN{978-1-4503-XXXX-X/18/06}

\begin{document}

\title{Retrieval-Augmented Generation Meets Data-Driven Tabula Rasa Approach for Temporal Knowledge Graph Forecasting}

\author{Geethan Sannidhi}
\affiliation{%
  \institution{IIIT}
  \country{Pune, India}
}
\email{geethansannidhi20@cse.iiitp.ac.in}

\author{Sagar Srinivas Sakhinana}
\affiliation{%
  \institution{TCS Research}
  \country{Bangalore, India}
}
\email{sagar.sakhinana@tcs.com}

\author{Venkataramana Runkana}
\affiliation{%
  \institution{TCS Research}
  \country{Pune, India}
}
\email{venkat.runkana@tcs.com}

\begin{abstract}
\vspace{-1mm}
Pre-trained large language models (PLLMs) like OpenAI ChatGPT and Google Gemini face challenges such as inaccurate factual recall, hallucinations, biases, and future data leakage for temporal Knowledge Graph (tKG) forecasting. To address these issues, we introduce \texttt{sLA-tKGF} (small-scale language assistant for tKG forecasting), which utilizes Retrieval-Augmented Generation (RAG) aided, custom-trained small-scale language models through a tabula rasa approach from scratch for effective tKG forecasting. Our framework constructs knowledge-infused prompts with relevant historical data from tKGs, web search results, and PLLMs-generated textual descriptions to understand historical entity relationships prior to the target time. It leverages these external knowledge-infused prompts for deeper understanding and reasoning of context-specific semantic and temporal information to zero-shot prompt small-scale language models for more accurate predictions of future events within tKGs. It reduces hallucinations and mitigates distributional shift challenges through comprehending changing trends over time. As a result, it enables more accurate and contextually grounded forecasts of future events while minimizing computational demands. Rigorous empirical studies demonstrate our framework's robustness, scalability, and state-of-the-art (SOTA) performance on benchmark datasets with interpretable and trustworthy tKG forecasting.
\vspace{-3mm}
\end{abstract}
 
\maketitle

\vspace{-3mm}
\section{\textbf{Introduction}}
\begin{figure}[ht!]
\centering
\resizebox{1.05\linewidth}{!}{ 
\hspace*{0mm}\includegraphics[keepaspectratio,height=5.0cm,trim=0.0cm 3.0cm 0cm 2.25cm,clip]{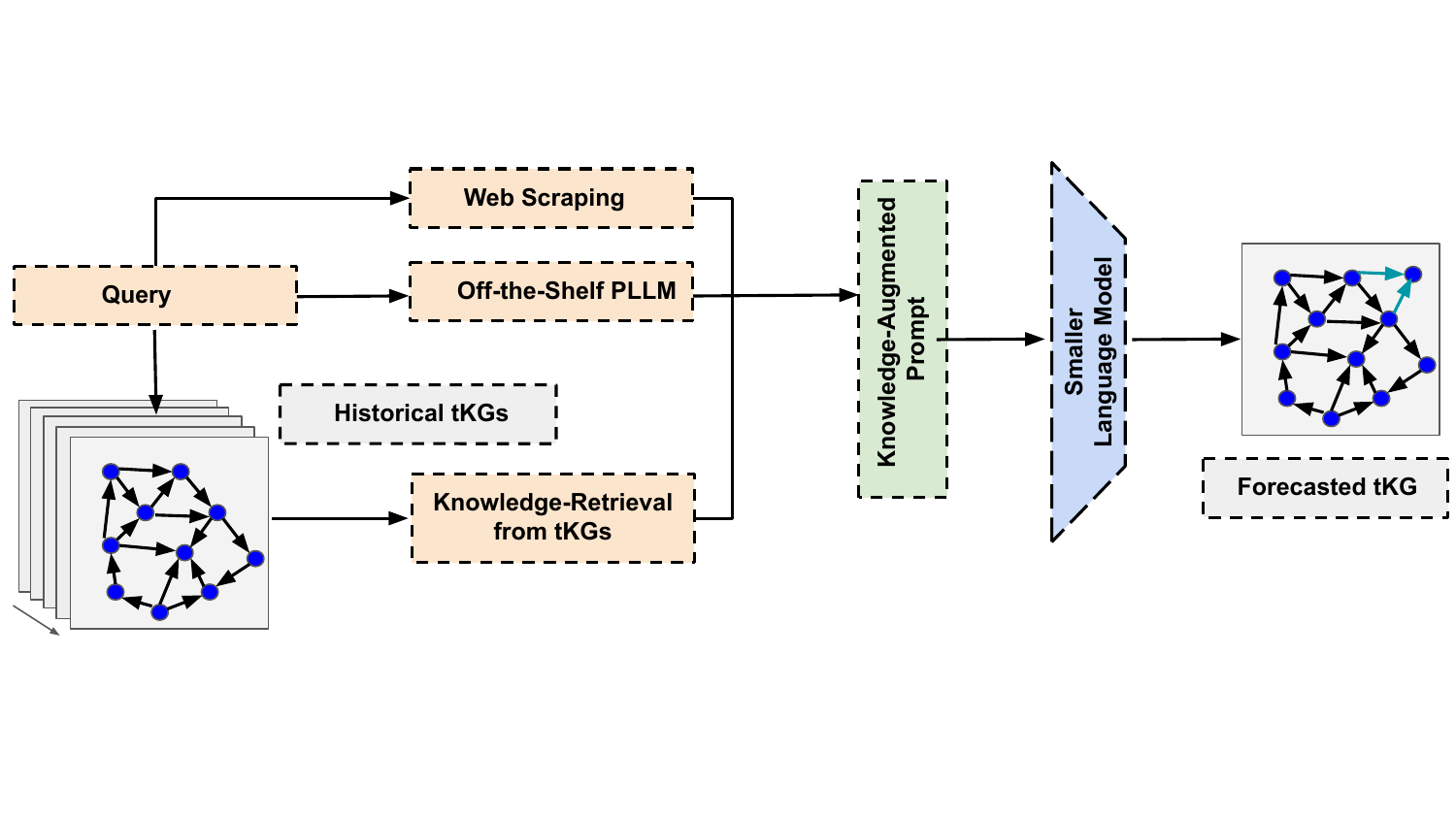} 
}
\vspace{-6mm}
\caption{The \texttt{sLA-tKGF} framework combines information retrieval from web scraping, historical knowledge from tKGs, and querying PLLMs to generate descriptions based on historical entity relationships to construct knowledge-augmented prompts for small-scale language models to achieve high factual accuracy and reliability in tKG forecasting.}
\label{fig:figure1}
\vspace{-4mm}
\end{figure}

\vspace{-1mm}
Knowledge Graphs (KGs) and their dynamic extension, Temporal Knowledge Graphs (tKGs), play a pivotal role in AI applications like recommendation engines and web searches by structuring data into graph-formatted databases. KGs utilize triples $(s, r, o)$—where $s$ is the subject, $o$ is the object, and $r$ describes their relation—to encode facts, while tKGs add a time element $(t)$ to represent time-validity of the fact, allowing them to capture how facts evolve over time. tKG reasoning, essential for deriving new insights, encompasses interpolation—to fill in historical data gaps~\cite{DBLP:conf/coling/JiangLGSCLS16, dasgupta2018hyte, DBLP:conf/aaai/GoelKBP20, DBLP:conf/emnlp/WuCCH20, lacroix2020tensor, garcia-duran-etal-2018-learning}—and extrapolation, for future event prediction~\cite{jin2019recurrent, trivedi2017knowevolve}. Predicting future events in tKGs is difficult, it requires handling unseen time periods and entirely new entities, demanding advanced methods to navigate the ever-changing nature of relationships. Closed-source pretrained large language models(PLLMs) like OpenAI ChatGPT\cite{achiam2023gpt} and Google Gemini\cite{team2023gemini} show potential for predicting future events due to their extensive pre-training knowledge and reasoning abilities. However, these large-scale models face challenges like fact recall issues stemming from complex architectures, resulting in unreliable predictions (hallucinations), potential bias from training data, and inadvertent leveraging of future data during pretraining, causing data leakage for tKG forecasting. Detecting data leakage is challenging when opaque training datasets are used for pretraining proprietary LLMs. Ensuring predictions rely solely on legitimate predictive abilities without the undue influence from future data remains a critical challenge for trustworthy tKG forecasting. To address the limitations of existing methods, we present \texttt{sLA-tKGF}, a small-scale language assistant for tKG forecasting based on Retrieval-Augmented Generation (RAG)~\cite{lewis2020retrieval} to ground predictions in historical context with source attribution and traceability. The framework employs a multi-layered stacked vanilla transformer architecture~\cite{vaswani2017attention} as its backbone language model and custom trained from scratch to avoid biases and data leakage inherent in pre-trained LLMs. The framework incorporates three key components: (i) retrieval of relevant historical knowledge from the tKGs. By retrieving historical facts based on context and semantic similarity to the query, we can infer causality and gain insights into temporal dynamics. Additionally, we employ text-embedding model and semantic similarity for query matching to filter out anachronistic/irrelevant information. (ii) utilizing PLLMs to analyze entity relationships prior to the target time to generate textual descriptions of historical entity relationships based on their internal knowledge acquired from vast pre-training text corpora. (iii) web scraping for up-to-date contextual information relevant to the query. We utilize advanced pre-processing techniques like sentence tokenization, temporal tagging, and date conversion for excluding future facts beyond the target time and retain appropriate scraped data. By incorporating information from diverse sources within a carefully crafted knowledge-augmented prompt, the framework generates factually accurate predictions grounded in historical context and ensures explainability. It minimizes the risk of bias, hallucinations, and avoids data leakage by pruning out-of-bound information. This `Tabula Rasa' approach of training the framework from scratch ensures a foundation of accountability and trustworthiness in tKG forecasting by ensuring predictions are truly based on past knowledge and patterns. Figure \ref{fig:figure1} provides an overview of the proposed approach. In summary, our proposed retrieval-augmented small-scale language model significantly improves tKG forecasting. It achieves this by dynamically accessing and leveraging external, continually evolving diverse data sources. This enhances the framework's utility in real-world applications, enabling it to generate historically accurate and well-explained forecasts. Experiments on real-world benchmark datasets demonstrate the framework effectiveness.

\vspace{-2mm}
\section{\textbf{Problem Formulation}}
\vspace{0mm}
An tKG captures the dynamic evolution of entity relationships over time, unlike static KGs, which are fixed. Let $\mathcal{V}$, $\mathcal{R}$, $\mathcal{T}$, and $\mathcal{F}$ symbolize sets of entities, relations, timestamps, and facts, respectively. A tKG snapshot $\mathcal{G}_{t_{i}} := \mathcal{V}_{i}, \mathcal{R}_{i}, \mathcal{F}_{i}$ at discrete timestamp $t_{i}$ offers a static view of graph relationships at that specific time. Each snapshot represents facts $\mathcal{F}_{i}$ as quadruples $q = (s, r, o, t_{i})$ at time $t_{i}$, indicating a timestamped relationship between entities $s$ and $o$ via relation $r$. A tKG comprises a sequence of these snapshots, $\mathcal{G} = {\mathcal{G}_{t_{1}}, \ldots, \mathcal{G}_{t_{\mathcal{T}}}}$, showcasing the graph's evolution over time. tKG forecasting predicts missing information in future snapshots based on past events. Given a target quadruple $(s_q, r_q, o_q, t_q)$, the goal is to predict the missing object entity $o_q$ in the query $(s_q, r_q, ?, t_q)$ using historical facts $\mathcal{O}_q = \{(s, r, o, t) \mid t < t_q\}$, which represent observed events before target time $t_q$. This process utilizes historical facts $\mathcal{C}_{(s_q, r_q, t_{q-1})}$ and $\mathcal{C}_{(s_q,t_{q-1})}$, representing past relationships involving $s_q$ and $r_q$, or $s_q$ with any relation, up to $t_{q-1}$, respectively. To forecast, all entities in $\mathcal{V}$ are considered potential candidates, and the most likely candidate is selected. Reciprocal relations are considered to account for bidirectional relationships, including both $(s, r, o, t)$ and its reciprocal $(o, r^{-1}, s, t)$, where $r^{-1}$ is the inverse of $r$. This approach ensures accurate predictions by considering the dynamics of relationships from both directions.

\vspace{-4.0mm}

\vspace{-2mm}
\section{\textbf{Proposed Method}} 
\texttt{sLA-tKGF} framework utilizes RAG to enhance the capabilities of the small-scale language model to predict future events in tKGs. We construct knowledge-infused prompts using historical tKG data, web information, and PLLM-generated past entity relationship descriptions for zero-shot prompting the language model for reliable and trustworthy forecasting, outperforming conventional methods.

\vspace{-2mm}
\subsection{\textbf{Web-Search-Driven Retrieval}}
\vspace{-1mm}
In developing a RAG framework that leverages web-based information, a critical challenge is addressed: excluding future facts before evaluating the relevance of retrieved web passages to a query. This approach focuses on preprocessing web passages to identify and exclude sentences containing future facts relative to a specified temporal threshold. This is achieved by using natural language processing techniques for sentence tokenization, temporal tagging, and resolving temporal expressions to absolute dates. By establishing a relevance period based on the query context, the framework filters out sentences with temporal references beyond this period, ensuring that only contextually and temporally relevant information is retained. Preparing the search data involves collecting various types of web content, parsing and chunking text, and embedding text chunks using text-embedding model. Each natural language question (verbalized query) is embedded similarly, then compared to the stored retrieved chunks for semantic similarity, and the most relevant chunks are integrated to generate an accurate response. By filtering out future or irrelevant information, our approach enhances the quality and reliability of real-time web information retrieval and generation.

\vspace{-3mm}
\subsection{\textbf{Retrieving Relevant Historical tKG Data}} 
\vspace{-1mm}
We leverage the following multi-step methods to incorporate relevant historical information from dynamic tKGs. (a) Query Verbalization: We convert structured queries $(s_q, r_q, ?, t_q)$ about future events into natural language questions using PLLMs like GPT-4. This allows the small-scale language model to interpret symbolic knowledge for forecasting tasks. For example, the query (\textit{Barack Obama}, \textit{visit}, ?, \textit{2015-01-25}) can be verbalized as  `\textit{Which country did Barack Obama visit on January 25, 2015?}. (b) Knowledge Retrieval \& Verbalization: Historical events relevant to the verbalized query up to the target time $t_q$ from evolving tKGs are incorporated to improve forecasting accuracy. It improves forecasting by grounding queries in historical context by identifying recurring patterns, trends, and causal relationships of the evolving entity relationships over time in graph-structured data. Motivated by the concept of window size in time series forecasting, which indicates the number of past observations used to predict future values, we adopt a similar approach in tKG forecasting. For a query at target time $t_q$ with subject $s_q$ and relation $r_q$, we retrieve `$m$' prior tKG snapshots $\mathcal{G}_{(t_q-m, t_q-1)}$ as background knowledge. A knowledge-infused prompt is constructed from historical facts related to $s_q$. Assuming a set of `$M$' retrieved facts related to the query, we describe this as follows:
 
\vspace{-2mm}
\begin{tcolorbox}[colback=gray!5!white,colframe=gray!75!black]
\vspace{-2mm}
\centering
\resizebox{1.2\linewidth}{!}{
\hspace{0mm}\begin{minipage}{1.45\linewidth}  
\begin{equation}
\hspace{-17mm}\mathcal{C}_{(s_q, r_q, ?, t_q)} = \left\{(s_i, r_i, o_i, t_i) \in \mathcal{C}_{(s_q, r_q, t_{q}-m, t_{q}-1)} \cup \mathcal{C}_{(s_q, t_{q}-m, t_{q}-1)} \hspace{0.75mm} \& \hspace{0.75mm} t < t_q \right\}_{i=1}^{\text{M}} \nonumber
\end{equation} 
\end{minipage}
}
\vspace{-3mm}
\end{tcolorbox}

\vspace{-1mm}
This set is constructed by combining facts involving $s_q$ within the time window $t_q - m$ to $t_q - 1$, covering both facts directly related to $s_q$ and $r_q$(i.e., $\mathcal{C}_{(s_q, t_{q}-m, t_{q}-1)}$), and those involving $s_q$ in any relation(i.e., $\mathcal{C}_{(s_q, r_q, t_{q}-m, t_{q}-1)}$), but only prior to $t_q$. This approach enhances forecasting accuracy by providing a comprehensive view of $s_q$'s historical occurrences. (c) Irrelevant Knowledge Rejection: We reject irrelevant facts using sentence embedding techniques, retaining only facts pertinent to the verbalized query based on semantic similarity. We then inject relevant historical knowledge into prompts to anchor predictions within historical tKG data, enhancing accuracy. For the query \( \text{(Barack Obama, visit, ?, 2015-01-25)} \) verbalized as `\textit{What country did Barack Obama visit on January 25, 2015?}', using relevant historical facts like `Barack Obama visited France on June 5, 2014', `Barack Obama visited Australia on November 15, 2014', etc., construct a knowledge-augmented prompt for the small-scale language model's zero-shot prediction on tKG forecasting.

\vspace{-3mm}
\subsection{\textbf{PLLMs-Generated Historical Summary}} 
\vspace{-1mm}
We query off-the-shelf PLLMs to generate a summary description of the historical entity relationships prior to the target time, based on the parametric knowledge acquired during pretraining.

\vspace{-3mm}
\subsection{Overall Framework}
\vspace{-1mm}
We construct knowledge-augmented prompts for grounded and interpretable forecasts by uniquely combining the relevant historical knowledge from tKGs, contemporary data through web search, and historical entity relationship descriptions generated by off-the-shelf PLLMs, setting a foundation for trustworthy tKG forecasting. In  summary, we perform memory augmentation for a small-scale language model trained from scratch by providing external knowledge through the RAG approach for more accurate tKG forecasts.

\vspace{-2mm}
\section{\textbf{Experiments and Results}}
\vspace{0mm}

\subsection{\textbf{Datasets}} 
\vspace{0mm}
We evaluate our framework with the following benchmark datasets: ICEWS14\cite{trivedi2017knowevolve}, ICEWS18\cite{DVN/28075_2015}, ICEWS05-15\cite{garcia-duran-etal-2018-learning}, YAGO\cite{mahdisoltani2013yago3}, and WIKI\cite{leblay2018deriving}. These diverse datasets, encapsulate real-world events in a quadruple format. The ICEWS datasets offers geo-coded events for systematic analysis and prediction relating to political violence, instability, and international relations from news reports over multi-decade time periods. For example, a quadruple fact $(s, r, o, t)$ such as (Barack Obama, visited, India, 2015-01-25). ICEWS14, ICEWS18, and ICEWS05-15 span events from 2014, january to october 2018, and 2005 to 2015, respectively. WIKI and YAGO, extracted from Wikipedia\cite{leblay2018deriving} and YAGO3-10\cite{mahdisoltani2013yago3} knowledge graphs (KGs), cover historical facts formatted as $(s, p, o, [T_s, T_e])$, with $T_s$ and $T_e$ marking start and end times. Our paper adopts the method from previous work\cite{jin2019recurrent}, for temporal fact discretization by breaking down multi-timestamp events into consecutive single events, improving accuracy by isolating each event. Table~\ref{tab:data} summarizes the dataset statistics. We utilize a new dataset from the Armed Conflict Location \& Event Data Project (ACLED)\footnote{\href{https://data.humdata.org/organization/acled}{https://data.humdata.org/organization/acled}}, detailing crisis situations globally. Our study targets hostility and aggression towards civilians in Cabo Delgado from January 1900 to March 2022, with the period from October 2021 to March 2022 as the test phase. Following the method in an earlier study\cite{jin2019recurrent}, we split datasets into training, validation, and test sets chronologically with an 80$\%$/10$\%$/10$\%$ ratio, except ICEWS14, split 50$\%$/50$\%$ due to no validation set. Splitting ensures training precedes validation and testing. Notably, many entities in the test sets are new compared to those in training and validation. Our research categorizes two types of tKG forecasting benchmarks: ICEWS datasets, featuring recurring short-lived events, e.g., $\textit{Xi Jinping visited the United States three times in 2015}$, and WIKI and YAGO datasets, capturing longer, non-recurring events, e.g., $\textit{Yossi Benayoun played for Chelsea FC from 2011 to 2013}$. 

\vspace{-2mm}
\subsection{\textbf{Prediction Settings}}
The tKG forecasting task has two primary prediction settings: (a) \texttt{Single Step}: It uses recent historical tKG data within a predefined window size(m) to predict at the next time point, incorporating facts up to time $t_{q-1}$ for predicting missing entities at time $t_q$. It utilizes actual historical data, not predictions from earlier points, to forecast at the next step. (b) \texttt{MultiStep}: It uses past framework predictions instead of ground truth facts for predicting missing entities at time $t_q$. These settings trade off computational cost, accuracy, and suitability for short-term or long-term forecasting.

\vspace{-2mm}
\subsection{\textbf{Baseline Methods}} 
Previous tKG forecasting research employed various methods for entity-relation modeling of historical events for missing entity prediction in future events. These methods included graph neural networks (GNNs)~\cite{jin-etal-2020-recurrent, li2021temporal, han-etal-2021-learning-neural, han2021explainable}, reinforcement learning~\cite{sun-etal-2021-timetraveler}, and logical rules~\cite{zhu2021learning, liu2022tlogic}. We benchmarked our tKG forecasting framework against established methods like Distmult~\cite{yang2014embedding}, TuckER~\cite{balazevic-etal-2019-tucker}, COMPGCN~\cite{vashishth2019composition}, TTransE~\cite{leblay2018deriving}, TA-Distmult~\cite{garcia-duran-etal-2018-learning}, CyGNet~\cite{zhu2020learning}, DE-SimplE~\cite{goel2020diachronic}, TNTComplEx~\cite{lacroix2020tensor}, and RE-Net~\cite{jin2019recurrent}. 
 
\subsection{\textbf{Evaluation Protocol}}
\label{sec: evaluation_protocol}
We evaluate our framework using standard metrics for the tKG forecasting task, specifically the Mean Reciprocal Rank (MRR) and the Hits@K metric. For each quadruple \((s_q, r_q, o_q, t_q)\) in the test set, we generate a query quadruple: \((s_q, r_q, ?, t_q)\) and the small-scale language model will rank all entities \(\mathcal{V}\) in its predictions. The rank of the actual missing entity \(o_q \in \mathcal{V}\) for each query quadruple is denoted by \(\psi(o_q)\). The MRR evaluation metric is defined as follows:

\vspace{-2mm}
\resizebox{0.925\linewidth}{!}{
\begin{minipage}{\linewidth}
\begin{align*}
\textit{MRR} = \frac{1}{|\mathcal{G}_{\text{test}}|} \sum_{q \in \mathcal{G}_{\text{test}}} \left(\frac{1}{\psi(o_q)}\right)
\end{align*}
\end{minipage}
}

MRR measures the average ranking of the actual missing entities in the predictions, with a higher MRR indicating better performance. The Hits@K metric (where \(K\) is in \{1, 3, 10\}) evaluates the accuracy of predicting missing entities at different positions in the predicted rankings. It measures the proportion of ground-truth missing entities ranked within the top-\(K\) positions, evaluating the framework's ability to predict missing entities accurately. For the tKG forecasting task, we consider  time-aware filtering as the evaluation setting to prevent valid predictions from being considered errors. For example, consider a test query like \(\textit{(Barack Obama, visit, ?, 2015)}\) with the true answer being \(\textit{India}\). Other valid predictions, such as \(\textit{(Barack Obama, visit, Germany, 2015)}\), might exist but would be removed by the time-aware filter. It allows us to determine the rank of \(\textit{India}\) without interference from alternative valid predictions.

\subsection{\textbf{Implementation Details}}
\label{sec:implementation_details}
By employing a `Tabula Rasa' approach—starting from a clean slate—the framework addresses the critical challenges of data leakage in predicting future events in tKGs.
Our orchestrated framework workflow utilizes LlamaIndex\cite{Liu_LlamaIndex_2022} for the development of PLLMs-powered applications. We utilize DuckDuckGo's search engine for searching the web. We access PLLMs using a text-based high-level API through Language Model as a Service (LaMaaS, \cite{sun2022black}). The framework is trained on tKG forecasting, aiming to minimize cross-entropy loss and predict missing entities in future events. Framework hyperparameters include a batch size ($b$) set at 48, the number of epochs ($e_{p}$) at 30, and the hidden/embedding dimension ($d$) at 128. The Adam optimizer (\cite{kingma2014adam}) is employed with an initial learning rate of $1e^{-3}$. A learning rate decay scheduler halves the learning rate if the validation loss doesn't improve over 5 epochs, and early stopping is implemented to prevent overfitting.
Four representative off-the-shelf PLLMs used are GPT-4, GPT-3.5-turbo, GPT-3.0-text-davinci-003, and Google Bard. Framework hyperparameters remain consistent across PLLMs and benchmark datasets, demonstrating versatility. The context window ($m$) is set to 25 for knowledge injection of historical events from evolving tKG into the prompt, and the quadruple retrieval hop is set to one. No limit is imposed on the number of facts retrieved for knowledge injection. Utilizing 8 V100 GPUs, each with 8 GB of GPU memory, ensures efficient training. Due to high computational costs, experiments are run three times, and averaged results  from independent experimental runs are reported in Tables \ref{tab:main-table1}, \ref{tab:main-table2}, and \ref{tab:main-table3} for robust comparisons. Refer appendix for hyperparameter tuning results.

\vspace{0mm}
\begin{table*}[htbp]
\centering
\resizebox{0.7\textwidth}{!}{
\hspace{0mm}\begin{tabular}{c c c c c c c c c} \hline
Dataset&\#Training&\#Validation&\#Test&\#Entities& \#Relation& Time granularity &$N_{\text{obs}}$\\ \hline
ICEWS14 \cite{trivedi2017knowevolve} & $323,895$ & $-$ & $341,409$ & $12,498$ & $260$ & 24 hours & $365$\\ 
ICEWS18 \cite{DVN/28075_2015} & $373,018$ & $45,995$ & $49,545$ & $23,033$ & $256$ & 24 hours & $304$\\ 
ICEWS05-15 \cite{garcia-duran-etal-2018-learning} & $369,104$ & $46,188$ & $46,037$ & $10,488$ & $251$ & 24 hours & $4,017$\\ 
WIKI \cite{leblay2018deriving} & $539,286$ & $67,538$ & $63,110$ & $12,554$ & $24$ & 1 year & $232$\\ 
YAGO \cite{mahdisoltani2013yago3} & $161,540$ & $19,523$ & $20,026$ & $10,623$ & $10$ & 1 year & $189$\\ 
ACLED-CD22 & 1,788 & 216 & 222 & 243 & 6& 24 hours & N/A \\
\hline
\end{tabular}}
\vspace{2mm}
\caption{The dataset statistics include the number of quadruples in the training, validation, and test sets, denoted as \#Training, \#Validation, and \#Test, respectively. $N_{\text{obs}}$ indicates the total snapshots of the temporal knowledge graph (tKG), with each snapshot capturing its state at a distinct time point.}
\label{tab:data}
\vspace{-8mm}
\end{table*}

\vspace{-3mm}
\subsection{\textbf{Results}}
\vspace{-1mm}
Our study compared the proposed framework with various supervised learning methods for tKG forecasting. Table \ref{tab:main-table1} demonstrates that \texttt{sLA-tKGF W/GPT-4}, outperformed baseline algorithms. We report baseline results from prior research \cite{lee2023temporal, gastinger2022evaluation} for fair and consistent comparison. Tables \ref{tab:main-table2} and \ref{tab:main-table3} provide further comparisons using several popular baselines, with results reported from an earlier study\cite{han-etal-2021-learning-neural}. Our experimental results confirm the effectiveness of \texttt{sLA-tKGF} framework in constructing knowledge-augmented prompts, involving: (1) retrieving historical knowledge from tKGs, (2) incorporating web search results for current information, and (3) using pre-trained large language models for summarizing historical entity-relationships, to query the small-scale language model and generate accurate and interpretable forecasts.

\vspace{-2mm}
\subsection{\textbf{Ablation Study}}
Given the complexity and multifaceted nature of the proposed \texttt{sLA-tKGF} framework for tKG forecasting, we propose several ablation experiments to evaluate the individual contributions of the components and their effectiveness within the framework. Each of these ablation studies can provide insights into the significance of the individual components and their collective impact on the overall efficacy of the \texttt{sLA-tKGF} framework. To design ablation studies for the \texttt{sLA-tKGF} framework, we systematically disable each component individually to evaluate its impact on the overall performance. This approach helps in understanding the contribution of individual components to the framework's accuracy, reliability, and robustness. The ablation study aims to evaluate the contributions of three key components in the framework for forecasting future events: (a) historical knowledge retrieval (HKR) from tKGs, (b) web search for contextual information (WSCI), and (c) descriptive text generation (DTG) using PLLMs. The HKR component, which retrieves relevant historical knowledge from dynamic tKGs, is assessed by contrasting its performance when incorporated versus omitted. Similarly, the impact of the WSCI component, providing up-to-date context prior to the target time through web search, is evaluated by disabling it and observing the changes in forecasting accuracy. The utility of the DTG component, which utilizes PLLMs to analyze historical entity relationships to generate descriptive texts, is quantified by including or excluding it and observing the impact on forecasting accuracy. The ablated variants are as follows:

\vspace{-1mm}
\begin{itemize}
\item \textbf{w/o HKR}: Omits historical knowledge retrieval from tKGs.
\item \textbf{w/o WSCI}: Omits web search for contextual information.
\item \textbf{w/o DTG}: Omits PLLMs for descriptive text generation.
\end{itemize}

\vspace{-1mm}
Table \ref{tab:main-table14} shows the ablation study results. The ablated variants  performance declines compared to the original framework (i.e., baseline for ablation study) and signifies the importance of the individual components. The study supports our approach of constructing knowledge-augmented prompts for querying the small-scale language model for improved forecasts.

\vspace{-2mm}
\subsection{\textbf{Deep Dive into Long-term tKG forecasting}}
Existing tKG forecasting research often overlooks insights from edge dissolution and formation. Our work considers the creation and dissolution of relationships (edges) between entities in tKGs as critical evolutionary factors for accurate tKG forecasting by capturing the essence of how relationships evolve and change over time. Nevertheless, the high dimensionality and dynamic complexity of tKGs pose challenges for both short and long-term event forecasting. Our approach uses historical context within an evolving tKG to improve both short-term and long-term forecasting accuracy. Short-term forecasts predict near-future changes, capturing the immediate evolution of events, while long-term forecasts anticipate broader trends  that unfold over an extended period. Standard tKG forecasting relies on static KG snapshots, denoted as $\mathcal{G}_{(t - t_m, t)}$, from a historical window $m$ up until time $t$, to predict events at time $t + \Delta t$ (e.g., one day). In real-world scenarios, the absence of graph information motivates the evaluation of forecasting techniques for making predictions about the distant future ($\Delta T \geq \Delta t$). In the context of long-term tKG forecasting, the objective is to predict missing entities at a future time $t + \Delta T$, beginning with an initial forecast period from $t - t_m$ to $t + \Delta t$. This initial forecast targets the short-term future, represented by $\Delta t$. The forecasting process then extends into the final forecast period, from $t + \Delta t$ to $t + \Delta T$, to focus on long-term predictions, where $\Delta T$ signifies the extended future. Experimental results on the ICEWS05-15 dataset, using \texttt{sLA-tKGF W/GPT-4}, for different $\Delta T$ values are shown in Table \ref{tab:main-table4}. As $\Delta T$ increases (1 day to 8 days), performance declines in both single-step and multi-step approaches, indicating increased difficulty in predicting further into the future. This decline in performance is attributed from the assumption that dynamics at $t + \Delta t$ apply at $t + \Delta T$. With larger $\Delta T$, this assumption falters due to data distributional shifts and the performance drops because of the lack of recent relevant historical facts in the augmented prompt for accurate predictions at $t + \Delta T$.

\vspace{-2mm}
\subsection{\textbf{Inductive Link Prediction in tKGs}}
In real-world applications, new entities emerge over time, necessitating a tKG forecasting framework that generalizes well to unseen data. An efficient framework must demonstrate robust generalization abilities to effectively handle new, unencountered entities. To evaluate this capability, we introduce the inductive link prediction task, which showcases the framework's ability to predict links involving unseen entities. This task involves selecting test

\begin{table*}[ht!]
   \begin{minipage}{0.975\textwidth} 
	\centering
	\small
	\resizebox{0.975\textwidth}{!}{
		\begin{tabular}{ccccccccccccccccc}
           \toprule
           \multirow{1}{*}{\textbf{Single-Step}} & \multicolumn{3}{c}{\textbf{YAGO}} & \multicolumn{3}{c}{\textbf{WIKI}} & \multicolumn{3}{c}{\textbf{ICEWS14}} & \multicolumn{3}{c}{\textbf{ICEWS18}} & \multicolumn{3}{c}{\textbf{ACLED-CD22}}\\
           \cmidrule(lr){2-4} \cmidrule(lr){5-7} \cmidrule(lr){8-10} \cmidrule(lr){11-13} \cmidrule(lr){14-16} & H@1 & H@3 & H@10 & H@1 & H@3 & H@10 & H@1 & H@3 & H@10 & H@1 & H@3 & H@10 & H@1 & H@3 & H@10 \\
           \midrule
           \texttt{RE-GCN}\cite{li2021temporal} & 0.787 & 0.842 & 0.884 & 0.747 & 0.817 & 0.846 & 0.313 & 0.473 & 0.626 & 0.223 &  0.367 & 0.525 &  0.446 &  0.545 &  0.608 \\
           \texttt{xERTE}\cite{han2021explainable} & 0.842 & 0.902 & 0.912 & 0.703 & 0.785 & 0.801 & 0.330 & 0.454 & 0.570 & 0.209 & 0.335 & 0.462 & 0.320 & 0.445 & 0.497 \\
           \texttt{TLogic}\cite{liu2022tlogic} & 0.740 & 0.789 & 0.791 &  0.786 &  0.860 & 0.870 & 0.332 & 0.476 & 0.602 & 0.204 & 0.336 & 0.480 & 0.009 & 0.045 & 0.094 \\
           \texttt{TANGO}\cite{han-etal-2021-learning-neural} & 0.590 & 0.646 & 0.677 & 0.483 & 0.514	& 0.527 & 0.272 & 0.408 & 0.550 & 0.191 & 0.318 & 0.462 & 0.327 & 0.482 & 0.599 \\
           \texttt{Timetraveler}\cite{sun-etal-2021-timetraveler} &  0.845 &  0.908 & 0.912 & 0.751 & 0.820 &	0.830 & 0.319 & 0.454 & 0.575 & 0.212 & 0.325 & 0.439 & 0.240 & 0.315 & 0.457\\
           \midrule
           \texttt{sLA-tKGF W/GPT-4} &   \bf 0.961 & \bf 0.994 & \bf 1.000 & \bf 0.895 & \bf 0.943 & \bf 0.994 & \bf 0.566 & \bf 0.689 & \bf 0.879 & \bf 0.608 & \bf 0.726 & \bf 0.854 & \bf 0.537 & \bf 0.689 & \bf 0.822 \\
           \texttt{sLA-tKGF W/GPT-3.5}  &  0.902 & 0.926 & 0.942 & 0.823 & 0.868 & 0.935 & 0.458 & 0.515 & 0.777 & 0.501 & 0.628 & 0.761 & 0.417 & 0.549 & 0.696\\
           \texttt{sLA-tKGF W/text-davinci-003} & 0.887 &  0.934 & 0.947 & 0.791 & 0.845 & 0.926 & 0.415 & 0.54 & 0.791 & 0.446 & 0.591 & 0.711 & 0.385 & 0.503 & 0.648\\
           \texttt{sLA-tKGF W/Google Bard} &  0.891 & 0.955 & 0.967 & 0.807 & 0.857 & 0.961 & 0.481 & 0.571 & 0.834 & 0.525 & 0.622 & 0.804 & 0.453 & 0.532 & 0.725\\
           \bottomrule
       \end{tabular}
	}
	\end{minipage}
	    \hfill
	\begin{minipage}{0.975\textwidth}
	\centering
	\small
	\resizebox{0.975\textwidth}{!}{
		\begin{tabular}{ccccccccccccccccc}
           \toprule
           \multirow{1}{*}{\textbf{Multi-Step}} & \multicolumn{3}{c}{\textbf{YAGO}} & \multicolumn{3}{c}{\textbf{WIKI}} & \multicolumn{3}{c}{\textbf{ICEWS14}} & \multicolumn{3}{c}{\textbf{ICEWS18}} & \multicolumn{3}{c}{\textbf{ACLED-CD22}}\\
           \cmidrule(lr){2-4} \cmidrule(lr){5-7} \cmidrule(lr){8-10} \cmidrule(lr){11-13} \cmidrule(lr){14-16} & H@1 & H@3 & H@10 & H@1 & H@3 & H@10 & H@1 & H@3 & H@10 & H@1 & H@3 & H@10 & H@1 & H@3 & H@10 \\
           \midrule
           \texttt{RE-GCN}\cite{li2021temporal} & 0.717 & 0.776 & 0.817 & 0.594 & 0.648 & 0.678 &  0.278 &  0.421 &  0.575 &  0.195 &  0.326 &  0.475 &  0.421 & 0.464 & 0.502 \\
           \texttt{RE-Net}\cite{jin-etal-2020-recurrent} & 0.534 & 0.613 & 0.662 & 0.472 & 0.507 & 0.530 & 0.278 & 0.408 & 0.549 & 0.184 & 0.314 & 0.461 & 0.238 & 0.445 & 0.563 \\
           \texttt{CyGNet}\cite{zhu2021learning} & 0.613 & 0.742 & 0.834 & 0.525 & 0.624 & 0.675 & 0.266 & 0.402 & 0.545 & 0.166 & 0.295 & 0.444 & 0.408 & 0.500 & 0.588 \\
           \texttt{TLogic}\cite{liu2022tlogic} & 0.631 & 0.706 & 0.715 & 0.613 &  0.663 & 0.682 & 0.265 & 0.395 & 0.531 & 0.155 & 0.272 & 0.412 & 0.009 & 0.045 & 0.094 \\
           \midrule
           \texttt{sLA-tKGF W/GPT-4} &  \bf 0.908 & \bf 0.938 & \bf 0.973 & \bf 0.846 & \bf 0.891 & \bf 0.968 & \bf 0.499 & \bf 0.633 & \bf 0.802 & \bf 0.545 & \bf 0.688 & \bf 0.828 & \bf 0.481 & \bf 0.648 & \bf 0.775 \\
           \texttt{sLA-tKGF W/GPT-3.5}  &  0.832 & 0.868 & 0.901 & 0.771 & 0.821 & 0.888 & 0.412 & 0.469 & 0.768 & 0.439 & 0.587 & 0.709 & 0.404 & 0.519 & 0.641\\
           \texttt{sLA-tKGF W/text-davinci-003} & 0.824 &  0.873 & 0.89 & 0.765 & 0.813 & 0.898 & 0.381 & 0.454 & 0.702 & 0.396 & 0.529 & 0.641 & 0.372 & 0.463 & 0.602\\
           \texttt{sLA-tKGF W/Google Bard} &  0.865 & 0.906 & 0.925 & 0.778 & 0.824 & 0.92 & 0.436 & 0.499 & 0.744 & 0.452 & 0.566 & 0.704 & 0.419 & 0.51 & 0.712\\
           \bottomrule
       \end{tabular}
	}
	\end{minipage}
   \vspace{1mm}
	\caption{The table shows the performance comparison of the proposed framework variants with various off-the-shelf PLLMs and baselines using the \textbf{Hits@K} metric for single-step and multi-step  tKG forecasting tasks. The model demonstrating the best performance for each dataset is highlighted in \textbf{bold}.}
	\label{tab:main-table1}
    \vspace{-0.5cm}
\end{table*} 

\vspace{-2mm}
\begin{table*}[ht!]
   \centering
   \resizebox{0.815\textwidth}{!}{
 \large\begin{tabular}{@{}|l|ccc|ccc|ccc|@{}}    
\toprule
       Datasets & \multicolumn{3}{|c}{\textbf{ICEWS05-15}} &  \multicolumn{3}{|c}{\textbf{ICEWS14}} & \multicolumn{3}{|c}{\textbf{ICEWS18}} \\
\midrule
       Model & Hits@1 & Hits@3 & Hits@10 & Hits@1 & Hits@3  & Hits@10 & Hits@1 & Hits@3  & Hits@10 \\
\midrule 
       \texttt{Distmult}\cite{yang2014embedding} & 16.10 & 27.67 & 42.42 & 8.15 & 15.31 & 27.66 & 9.68 & 18.12 & 31.21\\
       \texttt{TuckER}\cite{balazevic-etal-2019-tucker} & 17.01 & 29.93 & 47.81 & 11.23 & 20.77 & 33.94 & 12.58 & 22.60 & 37.27 \\

       \texttt{CompGCN}\cite{vashishth2019composition} & 20.72 & 32.51 & 47.87 & 10.12 & 19.49 & 33.11 & 12.01 & 22.96 & 38.15\\
\midrule
        \texttt{TTransE}\cite{leblay2018deriving} & 4.98 & 31.48 & 49.88 & 1.25 & 12.29 & 28.37 & 1.84 & 8.25 & 21.29 \\
      \texttt{TA-DistMult}\cite{garcia-duran-etal-2018-learning} & 14.77 & 27.80 & 44.22 & 4.72 & 10.54 & 21.48 & 5.58 & 12.04 & 22.82 \\
      
      \texttt{DE-SimplE} \cite{goel2020diachronic} & 26.33 & 39.41 & 53.97 & 13.77 & 23.68 & 37.15 & 11.53 & 21.86 & 34.80 \\
      
      \texttt{TNTComplEx}\cite{lacroix2020tensor} & 26.92 & 39.55 & 53.43 & 15.58 & 26.27 & 40.12 & 13.28 & 24.02 & 36.91 \\      

      \texttt{sLA-tKGF W/GPT-4} & \bf 55.52 & \bf 70.08 & \bf 94.66 & \bf 49.91 & \bf 63.35 & \bf 80.27 & \bf 54.58 & \bf 68.85 & \bf 82.86 \\   
      \texttt{sLA-tKGF W/GPT-3.5} & 45.74 & 53.89 & 73.76 & 41.27 & 46.93 & 76.86 & 43.97 & 58.76 & 70.92   \\   
      \texttt{sLA-tKGF text-davinci-003} & 43.87 & 49.27 & 65.78 & 38.14 & 45.43 & 70.26 & 39.64 & 52.97 & 64.18 \\   
      \texttt{sLA-tKGF W/Google Bard} & 50.13 & 63.51 & 83.73 & 43.64 & 49.96 & 74.49 & 45.27 & 56.62 & 70.43  \\   

\midrule 
   \end{tabular}
   }
   \vspace{1mm}
   \caption{The table shows the tKG forecasting results on three benchmark datasets on multi-step tKG forecasting task. The evaluation metrics include MRR ($\%$) and Hits@1/3/10 ($\%$). The best model for each dataset is highlighted in \textbf{bold}.}
   \label{tab:main-table2}
   \vspace{-4mm}
\end{table*}

\vspace{-2mm}
\begin{table*}[ht!]
   \centering
   \resizebox{0.475\textwidth}{!}{
   \large\begin{tabular}{@{}|l|ccc|ccc|@{}}
\toprule
       Datasets & \multicolumn{3}{|c}{\textbf{WIKI}}& \multicolumn{3}{|c|}{\textbf{YAGO}}\\
\midrule
       Model & Hits@1 & Hits@3  & Hits@10 & Hits@1 & Hits@3  & Hits@10\\
\midrule 
       \texttt{Distmult}\cite{yang2014embedding} & 46.17 & 52.81 & 54.13 & 47.39 & 59.81 & 68.52\\
       
       \texttt{TuckER}\cite{balazevic-etal-2019-tucker} & 46.12 & 53.60 & 54.86 & 47.42 & 59.63 & 68.96\\
       
       \texttt{CompGCN}\cite{vashishth2019composition} & 45.78 & 52.91 & 55.58 & 46.72 & 59.26 & 68.29\\
\midrule
        \texttt{TTransE}\cite{leblay2018deriving} & 21.67 & 34.43 & 42.39 & 18.12 & 40.91 & 51.21\\
        
      \texttt{TA-DistMult}\cite{garcia-duran-etal-2018-learning} & 39.92 & 48.73 & 51.71 & 48.15 & 59.61 & 66.71\\
      
      \texttt{DE-SimplE}\cite{goel2020diachronic} & 42.61 & 47.71 & 49.55 & 51.64 & 57.30 & 60.17 \\
      
      \texttt{TNTComplEx}\cite{lacroix2020tensor} & 40.04 & 49.31 & 52.03 & 52.92 & 61.33 & 66.69 \\       
      
      \texttt{W/GPT-4} & \bf 84.62 & \bf 89.17 & \bf 96.87 & \bf 90.85 & \bf 93.81 & \bf 97.36\\
      
      \texttt{W/GPT-3.5} & 77.14 & 82.15 &  88.87 & 83.26 & 86.82 & 90.18\\   
      
      \texttt{W/text-davinci-003} & 76.54 & 81.38 & 89.87 & 82.41 &  87.36 & 89.07\\ 
      \texttt{W/Google Bard} & 77.81 & 82.45 & 92.08 & 86.57 & 90.61 & 92.53\\   

\midrule
   \end{tabular}
   }
   \vspace{1mm}
   \small
   \caption{The table presents tKG forecasting results on two benchmark datasets on multi-step tKG forecasting task. The evaluation metrics are MRR ($\%$) and Hits@1/3/10 ($\%$). The best model for each dataset is highlighted in \textbf{bold}.}
\label{tab:main-table3} 
   \vspace{-5mm}
\end{table*}

\vspace{10mm}
quadruples where either the subject or object entity, or both, were not seen during training. For example, consider the ICEWS05-15 test set with the quadruple $\textit{(India, announce, new economic policy, 2015-09-01)}$  

\vspace{-2mm}
\begin{table*}[ht!]
   \begin{minipage}{1.0\textwidth}
	\centering
	\small
	\resizebox{1.0\textwidth}{!}{
		\begin{tabular}{ccccccccccccccccc}
           \toprule
           \multirow{1}{*}{\textbf{Single-Step}} & \multicolumn{3}{c}{\textbf{YAGO}} & \multicolumn{3}{c}{\textbf{WIKI}} & \multicolumn{3}{c}{\textbf{ICEWS14}} & \multicolumn{3}{c}{\textbf{ICEWS18}} & \multicolumn{3}{c}{\textbf{ACLED-CD22}}\\
           \cmidrule(lr){2-4} \cmidrule(lr){5-7} \cmidrule(lr){8-10} \cmidrule(lr){11-13} \cmidrule(lr){14-16} & H@1 & H@3 & H@10 & H@1 & H@3 & H@10 & H@1 & H@3 & H@10 & H@1 & H@3 & H@10 & H@1 & H@3 & H@10 \\
           \midrule
           \texttt{sLA-tKGF W/GPT-4} & \bf 0.961 & \bf 0.994 & \bf 1.000 & \bf 0.895 & \bf 0.943 & \bf 0.994 & \bf 0.566 & \bf 0.689 & \bf 0.879 & \bf 0.608 & \bf 0.726 & \bf 0.854 & \bf 0.537 & \bf 0.689 & \bf 0.822\\
           \texttt{sLA-tKGF w/o HKR}  & 0.773 & 0.798 & 0.804 & 0.729 & 0.771 & 0.807 & 0.462 & 0.553 & 0.707 & 0.490 & 0.591 & 0.684 & 0.431 & 0.564 & 0.671\\
           \texttt{sLA-tKGF w/o WSCI} & 0.827 & 0.857 & 0.864 & 0.770 & 0.828 & 0.868 & 0.498 & 0.596 & 0.773 & 0.530 & 0.622 & 0.742 & 0.467 & 0.590 & 0.704\\
           \texttt{sLA-tKGF w/o DTG}  & 0.866 & 0.913 & 0.911 & 0.819 & 0.857 & 0.895 & 0.516 & 0.633 & 0.791 & 0.549 & 0.655 & 0.782 & 0.485 & 0.633 & 0.742\\
           \bottomrule
       \end{tabular}
	}
	\end{minipage}
	\hfill
	    \begin{minipage}{1.0\textwidth}
	\centering
	\small
	\resizebox{1.0\textwidth}{!}{
		\begin{tabular}{ccccccccccccccccc}
           \toprule
           \multirow{1}{*}{\textbf{Multi-Step}} & \multicolumn{3}{c}{\textbf{YAGO}} & \multicolumn{3}{c}{\textbf{WIKI}} & \multicolumn{3}{c}{\textbf{ICEWS14}} & \multicolumn{3}{c}{\textbf{ICEWS18}} & \multicolumn{3}{c}{\textbf{ACLED-CD22}}\\
           \cmidrule(lr){2-4} \cmidrule(lr){5-7} \cmidrule(lr){8-10} \cmidrule(lr){11-13} \cmidrule(lr){14-16} & H@1 & H@3 & H@10 & H@1 & H@3 & H@10 & H@1 & H@3 & H@10 & H@1 & H@3 & H@10 & H@1 & H@3 & H@10 \\
           \midrule
           \texttt{sLA-tKGF W/GPT-4} &  \bf 0.908 & \bf 0.938 & \bf 0.973 & \bf 0.846 & \bf 0.891 & \bf 0.968 & \bf 0.499 & \bf 0.633 & \bf 0.802 & \bf 0.545 & \bf 0.688 & \bf 0.828 & \bf 0.481 & \bf 0.648 & \bf 0.775\\
           \texttt{sLA-tKGF w/o HKR}  & 0.742 & 0.752 & 0.792 & 0.685 & 0.727 & 0.782 & 0.402 & 0.510 & 0.651 & 0.443 & 0.562 & 0.664 & 0.386 & 0.526 & 0.629\\
           \texttt{sLA-tKGF w/o WSCI} & 0.788 & 0.820 & 0.848 & 0.734 & 0.761 & 0.827 & 0.433 & 0.550 & 0.693 & 0.466 & 0.587 & 0.716 & 0.422 & 0.568 & 0.664\\
           \texttt{sLA-tKGF w/o DTG}  & 0.834 & 0.853 & 0.885 & 0.765 & 0.811 & 0.876 & 0.457 & 0.576 & 0.727 & 0.494 & 0.620 & 0.753 & 0.442 & 0.588 & 0.701\\
           \bottomrule
       \end{tabular}
	}
	\end{minipage}
   \vspace{0.15cm}
	\caption{The table presents the experimental results from the ablation study conducted on tKG forecasting using benchmark datasets. The results show that the original framework consistently outperformed its ablated variants across all datasets. The ablation study supports the hypothesis of constructing knowledge-augmented prompts to improve the performance of the framework for optimal tkG forecasting. Nevertheless, all the ablated variants demonstrated a performance drop compared to the original framework, emphasizing the importance of the components that were disabled.}
	\label{tab:main-table14}
    \vspace{-0.4cm}
\end{table*}

\vspace{-1mm}
\begin{table*}[ht!]
    \begin{minipage}{0.975\textwidth}
	\centering
	\small
	\resizebox{0.975\textwidth}{!}{
		\begin{tabular}{ccccccccccccccccc}
            \toprule
            \multirow{1}{*}{\textbf{Single-Step}} & \multicolumn{3}{c}{\textbf{$\Delta T = \text{1 day}$}} & \multicolumn{3}{c}{\textbf{$\Delta T = \text{2 days}$}} & \multicolumn{3}{c}{\textbf{$\Delta T = \text{4 days}$}} & \multicolumn{3}{c}{\textbf{$\Delta T = \text{6 days}$}} & \multicolumn{3}{c}{\textbf{$\Delta T = \text{8 days}$}}\\
            \cmidrule(lr){2-4} \cmidrule(lr){5-7} \cmidrule(lr){8-10} \cmidrule(lr){11-13} \cmidrule(lr){14-16} & H@1 & H@3 & H@10 & H@1 & H@3 & H@10 & H@1 & H@3 & H@10 & H@1 & H@3 & H@10 & H@1 & H@3 & H@10 \\
            \midrule
            \texttt{sLA-tKGF W/GPT-4} & \bf 64.20 & \bf 79.78 & \bf 99.52 & 60.72 & 74.37 & 96.18 &  57.34 &  74.11 &  88.00 &  56.64 &  70.48 &  89.16 &  52.49 &  65.43 &  85.69 
  \\
            \bottomrule
        \end{tabular}
	}
	\end{minipage}
	\hfill
	    \begin{minipage}{0.975\textwidth}
	\centering
	\small
	\resizebox{0.975\textwidth}{!}{
		\begin{tabular}{ccccccccccccccccc}
            \toprule
            \multirow{1}{*}{\textbf{Multi-Step}} & \multicolumn{3}{c}{\textbf{$\Delta T = \text{1 day}$}} & \multicolumn{3}{c}{\textbf{$\Delta T = \text{2 days}$}} & \multicolumn{3}{c}{\textbf{$\Delta T = \text{4 days}$}} & \multicolumn{3}{c}{\textbf{$\Delta T = \text{6 days}$}} & \multicolumn{3}{c}{\textbf{$\Delta T = \text{8 days}$}}\\
            \cmidrule(lr){2-4} \cmidrule(lr){5-7} \cmidrule(lr){8-10} \cmidrule(lr){11-13} \cmidrule(lr){14-16} & H@1 & H@3 & H@10 & H@1 & H@3 & H@10 & H@1 & H@3 & H@10 & H@1 & H@3 & H@10 & H@1 & H@3 & H@10 \\
            \midrule
            \texttt{sLA-tKGF W/GPT-4} & \bf 55.52 & \bf 70.08 & \bf 94.66 & 51.48 &  66.22 &  92.38 &  46.21 &  58.89 &  86.37 &  40.47 &  50.60 &  77.21 &  33.69 &  43.92 &  63.70 \\
            \bottomrule
        \end{tabular}
	}
	\end{minipage}
    \vspace{0.2cm}
	\caption{The table presents the results for the long-term tKG forecasting task on the ICEWS05-15 dataset across different forecast horizon($\Delta T$) in terms of the Mean Reciprocal Rank(MRR) metric.}
	\label{tab:main-table4}
     \vspace{-0.1cm}
\end{table*}

\vspace{0mm}
\begin{table}[ht!]
    \begin{minipage}{0.625\textwidth}
	\centering
	\small
	\resizebox{0.55\textwidth}{!}{
		\hspace{-20mm}\begin{tabular}{ccccccccccccccccc}
            \toprule
            \multirow{1}{*}{\textbf{Single-Step}} & \multicolumn{3}{c}{\textbf{ICEWS05-15}} & \\
            \cmidrule(lr){2-4} \cmidrule(lr){5-7} \cmidrule(lr){8-10} \cmidrule(lr){11-13} \cmidrule(lr){14-16} & H@1 & H@3 & H@10 & \\
            \midrule
            \texttt{sLA-tKGF W/GPT-4 (ILP)} & \bf 57.50 & \bf 73.01 & \bf 93.99 \\
            \texttt{sLA-tKGF W/GPT-3.5 (ILP)} & 50.89 & 63.92 & 80.20 \\         
            \texttt{sLA-tKGF text-davinci-003 (ILP)} & 49.98 & 62.23 & 84.04 \\  
            \texttt{sLA-tKGF W/Google Bard (ILP)} & 51.96 & 68.88 & 86.22 \\ 
            \bottomrule
        \end{tabular}
	}
	\end{minipage}
	\hfill
	    \begin{minipage}{0.625\textwidth}
	\centering
	\small
	\resizebox{0.55\textwidth}{!}{
		\hspace{-20mm}\begin{tabular}{ccccccccccccccccc}
            \toprule
            \multirow{1}{*}{\textbf{Multi-Step}} & \multicolumn{3}{c}{\textbf{ICEWS05-15}} & \\
            \cmidrule(lr){2-4} \cmidrule(lr){5-7} \cmidrule(lr){8-10} \cmidrule(lr){11-13} \cmidrule(lr){14-16} & H@1 & H@3 & H@10 & \\
            \midrule
            \texttt{sLA-tKGF W/GPT-4 (ILP)} & \bf 48.27 & \bf 62.92 & \bf 86.46 \\
            \texttt{sLA-tKGF W/GPT-3.5 (ILP)} & 41.16 & 54.96 & 76.65 \\
            \texttt{sLA-tKGF text-davinci-003 (ILP)} & 41.94 & 53.99 & 76.96 \\
            \texttt{sLA-tKGF W/Google Bard (ILP)} & 44.56 & 57.03 & 81.34 \\
            \bottomrule
        \end{tabular}
	}
	\end{minipage}
    \vspace{0.1cm}
	\caption{The table presents the results for inductive future link prediction task on ICEWS05-15 dataset in terms of Mean Reciprocal Rank(MRR) metric.}
	\label{tab:main-table5}
     \vspace{-1.0cm}
\end{table}

Since the framework hasn't encountered $\textit{new economic policy}$ during training, this serves as an instance of an unobserved entity.  We call this evaluation, `inductive link prediction analysis'. We conducted future link prediction experiments on sets of inductive link prediction quadruples, and the results are shown in Table \ref{tab:main-table5}. We evaluated our framework against robust baselines, using the ICEWS05-15 dataset. Despite not being explicitly designed for knowledge graph tasks like inductive link prediction, our proposed framework (\texttt{sLA-tKGF W/GPT-4}) surprisingly outperforms all baselines across various metrics, as demonstrated in Table \ref{tab:main-table5}. This success stems from our framework's ability to leverage both the web search results and historical entity-relationships based descriptions generated by PLLMs using implicit pre-training knowledge. In essence, when applied to the test dataset, the framework utilizes patterns and information from its training data, augmented by knowledge-rich prompts at inference time, to successfully predict relationships between unseen entities within an evolving tKG.

\section{\textbf{Conclusion}}
Despite their strengths, pre-trained large language models struggle with future predictions for tKG forecasting due to limitations including hallucinations, inaccurate fact recall, and future data leakage. Our framework tackles these challenges by utilizing a retrieval-augmented small-scale language model, trained from a clean-slate approach with knowledge-augmented prompts to achieve accurate tKG forecasts. We construct these prompts that include historical data from tKGs, web search results, and text descriptions generated by off-the-shelf PLLMs, enabling contextually grounded forecasts with improved factual accuracy, reducing hallucinations and overcoming distributional shifts. This scalable, robust approach achieves state-of-the-art (SOTA) performance on benchmark tKG datasets. It paves the way for trustworthy tKG forecasting by offering traceability, explainability, and interpretability.

\clearpage
\newpage

\bibliographystyle{ACM-Reference-Format}
\bibliography{sample-base}

\clearpage
\newpage

\section{Additional Experiments}

\subsection{\textbf{The Impact of Temporal and Sequential Information on the Zero-Shot tKG Forecasting Task}}
The proposed framework, \texttt{sLA-tKGF}, introduces a novel two-pronged approach to predict future events in tKGs by combining a clean-slate trained, small-scale language model with the Retrieval-Augmented Generation (RAG) technique. It emphasizes accuracy and bias-free predictions by leveraging historical tKG data, current web information, and contextually relevant historical entity relationship descriptions generated by PLLMs. We examined the performance of the \texttt{sLA-tKGF} framework by understanding temporal information in retrieved historical events from tKGs, comparing its performance in the tKG forecasting task with and without explicit timestamps in the knowledge-augmented prompt. We also investigated the impact of shuffling historical facts without time information on the performance of the \texttt{sLA-tKGF} framework. The results, shown in Table \ref{tab:main-table6a}, reveal that the \texttt{sLA-tKGF} framework's performance worsens without temporal information. This decline is exacerbated by the random arrangement of events, as demonstrated in Table \ref{tab:main-table6b}. These outcomes underscore the framework's dependence on chronological sequencing for accurate predictions, emphasizing the critical role of temporal information and sequence in the accuracy of the \texttt{sLA-tKGF W/GPT-4} framework for tKG forecasting. The consistent significance of temporal and sequence information across datasets (YAGO, WIKI, ICEWS14, ICEWS18, ACLED-CD22) reinforces the reliability and applicability of these findings.

\subsection{\textbf{Effect of Types of Knowledge-Infused Augmented Prompts on tKG Forecasting}}
In this section, we evaluate the impact of augmenting natural language questions(verbalized queries) with external knowledge from historical events sampled from tKGs on the \texttt{sLA-tKGF W/GPT-4} framework's performance in tKG forecasting tasks. We explore various knowledge retrieval strategies for constructing knowledge-augmented prompts, focusing on optimizing the framework's performance in tKG forecasting. Table \ref{tab:main-table7} shows how different strategies affect the \texttt{sLA-tKGF W/GPT-4} framework's performance across datasets (YAGO, WIKI, ICEWS14, ICEWS18, ACLED-CD22) for Single-Step and Multi-Step forecasting. The results demonstrate that selectively incorporating relevant prior knowledge significantly enhances tKG forecasting performance, surpassing baselines that either omit additional knowledge or use it indiscriminately.

\begin{itemize}
\item The $\textbf{No Knowledge (NK)}$ baseline applies the given query without external knowledge.
\item The $\textbf{Random Knowledge (RK)}$ baseline constructs prompts with randomly selected historical facts.
\item The $\textbf{Popular Knowledge (PK)}$ baseline builds prompts using the most common relationship-based historical facts from the tKG.
\item The $\textbf{sLA-tKGF (Ours)}$ framework selects the most pertinent historical facts for the query, excludes irrelevant facts, and uses this filtered knowledge for forecasting.
\end{itemize}

Our experiments show that even randomly included historical facts improve performance compared to not using any additional knowledge. However, knowledge extracted specifically from historical events within the tKGs proves even more effective than other knowledge retrieval strategies. This finding underscores the potential of retrieving relevant historical facts to provide pertinent information for tKG forecasting tasks.

\subsection{\textbf{Balancing Accuracy and Efficiency: Optimizing Historical Context in Augmented Prompts for tKG Forecasting}}
Our framework utilizes historical events or facts from time $(t_q - m)$ to $t_q$ to predict missing entities in the query quadruple at $t_q$, where $m$ represents the historical context window size, a configurable hyperparameter. We experimented with various lengths to evaluate the impact of historical context on the forecasting performance of the \texttt{sLA-tKGF W/GPT-4} framework for tKGs. According to Table \ref{tab:main-table8}, leveraging a greater amount of past data to generate knowledge-augmented prompts for the small-scale language model enhances the accuracy of missing entity predictions, as evidenced by improved mean reciprocal rank (MRR) scores, albeit at the expense of higher computational demands. The small-scale language model within the \texttt{sLA-tKGF W/GPT-4} framework is constrained by the maximum input token sequence length. Although extending the historical window marginally increases MRR scores, it leads to the creation of longer and more complex prompts that are impractical for broad-scale application. To strike a balance between accuracy and computational efficiency, we selected a historical context window of 25 for our experiments. We varied the historical facts in the prompts to identify the optimal setup for generating accurate forecasts, with the goal of minimizing wall-clock time. Wall-clock time for \texttt{sLA-tKGF W/GPT-4} queries—the time elapsed from query submission to response—is influenced by the size of the small-scale language model, the complexity of the query, and the available computational resources. While typically, small-scale language models return responses within seconds, more complex or extensive queries may require longer processing times.

\begin{table*}[ht!]
    \begin{minipage}{1.0\textwidth}
	\centering
	\small
	\resizebox{1.0\textwidth}{!}{
		\begin{tabular}{ccccccccccccccccc}
            \toprule
            \multirow{1}{*}{\textbf{Single-Step}} & \multicolumn{3}{c}{\textbf{YAGO}} & \multicolumn{3}{c}{\textbf{WIKI}} & \multicolumn{3}{c}{\textbf{ICEWS14}} & \multicolumn{3}{c}{\textbf{ICEWS18}} & \multicolumn{3}{c}{\textbf{ACLED-CD22}}\\
            \cmidrule(lr){2-4} \cmidrule(lr){5-7} \cmidrule(lr){8-10} \cmidrule(lr){11-13} \cmidrule(lr){14-16} & H@1 & H@3 & H@10 & H@1 & H@3 & H@10 & H@1 & H@3 & H@10 & H@1 & H@3 & H@10 & H@1 & H@3 & H@10 \\
            \midrule
\texttt{sLA-tKGF W/GPT-4} & \bf 0.961 & \bf 0.994 & \bf 1.000 & \bf 0.895 & \bf 0.943 & \bf 0.994 & \bf 0.566 & \bf 0.689 & \bf 0.879 & \bf 0.608 & \bf 0.726 & \bf 0.854 & \bf 0.537 & \bf 0.689 & \bf 0.822 \\
\texttt{sLA-tKGF W/GPT-4 W/o TS} &  0.832 &  0.903 &  0.926 &  0.781 &  0.845 &  0.882 &  0.516 &  0.593 &  0.799 &  0.547 &  0.638 &  0.750 &  0.466 &  0.629 &  0.725 \\
            \bottomrule
        \end{tabular}
	}
	\end{minipage}
	\hfill
	    \begin{minipage}{1.0\textwidth}
	\centering
	\small
	\resizebox{1.0\textwidth}{!}{
		\begin{tabular}{ccccccccccccccccc}
            \toprule
            \multirow{1}{*}{\textbf{Multiple-Step}} & \multicolumn{3}{c}{\textbf{YAGO}} & \multicolumn{3}{c}{\textbf{WIKI}} & \multicolumn{3}{c}{\textbf{ICEWS14}} & \multicolumn{3}{c}{\textbf{ICEWS18}} & \multicolumn{3}{c}{\textbf{ACLED-CD22}}\\
            \cmidrule(lr){2-4} \cmidrule(lr){5-7} \cmidrule(lr){8-10} \cmidrule(lr){11-13} \cmidrule(lr){14-16} & H@1 & H@3 & H@10 & H@1 & H@3 & H@10 & H@1 & H@3 & H@10 & H@1 & H@3 & H@10 & H@1 & H@3 & H@10 \\
            \midrule
\texttt{sLA-tKGF W/GPT-4} & \bf 0.908 & \bf 0.938 & \bf 0.973 & \bf 0.846 & \bf 0.891 & \bf 0.968 & \bf 0.499 & \bf 0.633 & \bf 0.802 & \bf 0.545 & \bf 0.688 & \bf 0.828 & \bf 0.481 & \bf 0.648 & \bf 0.775 \\
\texttt{sLA-tKGF W/GPT-4 W/o TS} &  0.752 &  0.839 &  0.864 &  0.751 &  0.764 &  0.837 &  0.415 &  0.519 &  0.658 &  0.476 &  0.606 &  0.707 &  0.397 &  0.570 &  0.691 \\
            \bottomrule
        \end{tabular}
	}
	\end{minipage}
    \vspace{0.1cm}
	\caption{The table shows the performance of the \texttt{sLA-tKGF W/GPT-4} framework on tKG forecasting declines without timestamped (TS) facts. This suggests that \texttt{sLA-tKGF W/GPT-4} can leverage temporal information in historical facts to improve its performance on tKG forecasting tasks.}
	\label{tab:main-table6a}
     \vspace{-0.2cm}
\end{table*}

\begin{table*}[ht!]
    \begin{minipage}{1.0\textwidth}
	\centering
	\small
	\resizebox{1.0\textwidth}{!}{
		\begin{tabular}{ccccccccccccccccc}
            \toprule
            \multirow{1}{*}{\textbf{Single-Step}} & \multicolumn{3}{c}{\textbf{YAGO}} & \multicolumn{3}{c}{\textbf{WIKI}} & \multicolumn{3}{c}{\textbf{ICEWS14}} & \multicolumn{3}{c}{\textbf{ICEWS18}} & \multicolumn{3}{c}{\textbf{ACLED-CD22}}\\
            \cmidrule(lr){2-4} \cmidrule(lr){5-7} \cmidrule(lr){8-10} \cmidrule(lr){11-13} \cmidrule(lr){14-16} & H@1 & H@3 & H@10 & H@1 & H@3 & H@10 & H@1 & H@3 & H@10 & H@1 & H@3 & H@10 & H@1 & H@3 & H@10 \\
            \midrule
\texttt{sLA-tKGF W/GPT-4} & \bf 0.961 & \bf 0.994 & \bf 1.000 & \bf 0.895 & \bf 0.943 & \bf 0.994 & \bf 0.566 & \bf 0.689 & \bf 0.879 & \bf 0.608 & \bf 0.726 & \bf 0.854 & \bf 0.537 & \bf 0.689 & \bf 0.822 \\
\texttt{sLA-tKGF W/GPT-4 W/o TS W/RS} &  0.784 &  0.803 &  0.848 &  0.744 &  0.773 &  0.799 &  0.460 &  0.579 &  0.727 &  0.510 &  0.593 &  0.697 &  0.443 &  0.554 &  0.686  \\
            \bottomrule
        \end{tabular}
	}
	\end{minipage}
	\hfill
	    \begin{minipage}{1.0\textwidth}
	\centering
	\small
	\resizebox{1.0\textwidth}{!}{
		\begin{tabular}{ccccccccccccccccc}
            \toprule
            \multirow{1}{*}{\textbf{Multiple-Step}} & \multicolumn{3}{c}{\textbf{YAGO}} & \multicolumn{3}{c}{\textbf{WIKI}} & \multicolumn{3}{c}{\textbf{ICEWS14}} & \multicolumn{3}{c}{\textbf{ICEWS18}} & \multicolumn{3}{c}{\textbf{ACLED-CD22}}\\
            \cmidrule(lr){2-4} \cmidrule(lr){5-7} \cmidrule(lr){8-10} \cmidrule(lr){11-13} \cmidrule(lr){14-16} & H@1 & H@3 & H@10 & H@1 & H@3 & H@10 & H@1 & H@3 & H@10 & H@1 & H@3 & H@10 & H@1 & H@3 & H@10 \\
            \midrule
\texttt{sLA-tKGF W/GPT-4} & \bf 0.908 & \bf 0.938 & \bf 0.973 & \bf 0.846 & \bf 0.891 & \bf 0.968 & \bf 0.499 & \bf 0.633 & \bf 0.802 & \bf 0.545 & \bf 0.688 & \bf 0.828 & \bf 0.481 & \bf 0.648 & \bf 0.775 \\
\texttt{sLA-tKGF W/GPT-4 W/o TS W/RS} &  0.728 &  0.748 &  0.767 &  0.673 &  0.703 &  0.760 &  0.399 &  0.503 &  0.631 &  0.438 &  0.547 &  0.660 &  0.381 &  0.511 &  0.610 \\
            \bottomrule
        \end{tabular}
	}
	\end{minipage}
    \vspace{0.1cm}
	\caption{The table shows performance of \texttt{sLA-tKGF W/GPT-4} framework on tKG forecasting declines when the timestamp removed facts in the augmented prompt are randomly shuffled(RS). This suggests that \texttt{sLA-tKGF W/GPT-4} can leverage the chronological order of events to improve performance on tKG forecasting tasks.}
	\label{tab:main-table6b}
     \vspace{-0.1cm}
\end{table*}

\vspace{-3mm}
\begin{table*}[ht!]
    \begin{minipage}{1.0\textwidth}
	\centering
	\small
	\resizebox{1.0\textwidth}{!}{
		\begin{tabular}{ccccccccccccccccc}
            \toprule
            \multirow{1}{*}{\textbf{Single-Step}} & \multicolumn{3}{c}{\textbf{YAGO}} & \multicolumn{3}{c}{\textbf{WIKI}} & \multicolumn{3}{c}{\textbf{ICEWS14}} & \multicolumn{3}{c}{\textbf{ICEWS18}} & \multicolumn{3}{c}{\textbf{ACLED-CD22}}\\
            \cmidrule(lr){2-4} \cmidrule(lr){5-7} \cmidrule(lr){8-10} \cmidrule(lr){11-13} \cmidrule(lr){14-16} & H@1 & H@3 & H@10 & H@1 & H@3 & H@10 & H@1 & H@3 & H@10 & H@1 & H@3 & H@10 & H@1 & H@3 & H@10 \\
            \midrule
\texttt{sLA-tKGF W/GPT-4 (Ours)} & \bf 0.961 & \bf 0.994 & \bf 1.000 & \bf 0.895 & \bf 0.943 & \bf 0.994 & \bf 0.566 & \bf 0.689 & \bf 0.879 & \bf 0.608 & \bf 0.726 & \bf 0.854 & \bf 0.537 & \bf 0.689 & \bf 0.822 \\
\texttt{sLA-tKGF W/GPT-4 W/NK}  &  0.765 &  0.835 &  0.855 &  0.740 &  0.775 &  0.857 &  0.427 &  0.549 &  0.712 &  0.507 &  0.555 &  0.712 &  0.414 &  0.520 &   0.664 \\
\texttt{sLA-tKGF W/GPT-4 W/RK} &  0.789 &  0.858 &  0.885 &  0.757 &  0.804 &  0.887 &  0.439 &  0.567 &  0.738 &  0.527 &  0.571 &  0.742 &  0.429 &  0.539 &   0.683 \\
\texttt{sLA-tKGF W/GPT-4 W/PK} &  0.815 &  0.870 &  0.912 &  0.780 &  0.838 &  0.924 &  0.458 &  0.590 &  0.765 &  0.548 &  0.596 &  0.775 &  0.445 &  0.574 &   0.705 \\
            \bottomrule
        \end{tabular}
	}
	\end{minipage}
	\hfill
	    \begin{minipage}{1.0\textwidth}
	\centering
	\small
	\resizebox{1.0\textwidth}{!}{
		\begin{tabular}{ccccccccccccccccc}
            \toprule
            \multirow{1}{*}{\textbf{Multi-Step}} & \multicolumn{3}{c}{\textbf{YAGO}} & \multicolumn{3}{c}{\textbf{WIKI}} & \multicolumn{3}{c}{\textbf{ICEWS14}} & \multicolumn{3}{c}{\textbf{ICEWS18}} & \multicolumn{3}{c}{\textbf{ACLED-CD22}}\\
            \cmidrule(lr){2-4} \cmidrule(lr){5-7} \cmidrule(lr){8-10} \cmidrule(lr){11-13} \cmidrule(lr){14-16} & H@1 & H@3 & H@10 & H@1 & H@3 & H@10 & H@1 & H@3 & H@10 & H@1 & H@3 & H@10 & H@1 & H@3 & H@10 \\
            \midrule
\texttt{sLA-tKGF W/GPT-4 (Ours)} & \bf 0.908 & \bf 0.938 & \bf 0.973 & \bf 0.846 & \bf 0.891 & \bf 0.968 & \bf 0.499 & \bf 0.633 & \bf 0.802 & \bf 0.545 & \bf 0.688 & \bf 0.828 & \bf 0.481 & \bf 0.648 & \bf 0.775 \\
\texttt{sLA-tKGF W/GPT-4 W/NK} & 0.733 & 0.774 & 0.844 & 0.722 & 0.769 & 0.854 & 0.401 & 0.580 & 0.713 & 0.468 & 0.642 & 0.738 & 0.413 & 0.575 & 0.692 \\
\texttt{sLA-tKGF W/GPT-4 W/RK} & 0.791 & 0.818 & 0.898 & 0.777 & 0.803 & 0.901 & 0.429 & 0.615 & 0.752 & 0.500 & 0.679 & 0.794 & 0.440 & 0.616 & 0.730 \\
\texttt{sLA-tKGF W/GPT-4 W/PK} & 0.819 & 0.848 & 0.930 & 0.803 & 0.855 & 0.944 & 0.445 & 0.635 & 0.795 & 0.521 & 0.715 & 0.819 & 0.459 & 0.641 & 0.759 \\
            \bottomrule
        \end{tabular}
	}
	\end{minipage}
    \vspace{0.1cm}
	\caption{The table displays the results of the study investigating the impact of various types of knowledge-infused augmented prompts on the performance of the \texttt{sLA-tKGF W/GPT-4} framework in tKG forecasting, assessed using benchmark datasets.}
	\label{tab:main-table7}
     \vspace{-0.1cm}
\end{table*}

\subsection{\textbf{Study of tKG forecasting task with non-uniform time intervals}}
Many state-of-the-art techniques struggle with tKGs featuring irregular time intervals, unlike the \texttt{sLA-tKGF} framework, which effectively addresses this issue by leveraging knowledge-augmented prompting for small-scale language models in temporal KG forecasting. The framework excels in handling real-world complexities and data sparsity, capturing complex dynamics and causal relationships more accurately, thus offering a versatile and reliable solution for temporal KG forecasting. Experimental evaluations on the ICEWS05-15$\_$continuous dataset, a subset created by sampling from the original ICEWS05-15 dataset to simulate non-periodic observations in continuous time with 1-4 units interval, support our claim. The \texttt{sLA-tKGF} framework, trained on this benchmark and assessed using the Mean Reciprocal Rank (MRR) metric, demonstrates strong performance, especially with the \texttt{sLA-tKGF W/GPT-4} configuration, on temporal KGs with irregular intervals. The dataset statistics for ICEWS05-15$\_\text{continuous}$ are presented in Table \ref{tab:main-table10}. We trained the \texttt{sLA-tKGF} framework with various off-the-shelf Pretrained Large Language Models (PLLMs) on this new benchmark dataset and evaluated their performance using the Mean Reciprocal Rank (MRR) metric. As demonstrated in Table \ref{tab:main-table9}, our results validate that the \texttt{sLA-tKGF W/GPT-4} framework, exhibits strong performance on tKG forecasting task with irregular time intervals.

\vspace{-3mm}
\begin{table*}[ht!]
    \begin{minipage}{1.0\textwidth}
	\centering
	\small
	 \resizebox{1.0\textwidth}{!}{
		\begin{tabular}{ccccccccccccccccc}
            \toprule
            \multirow{1}{*}{\textbf{Single-Step}} & \multicolumn{3}{c}{\textbf{YAGO}} & \multicolumn{3}{c}{\textbf{WIKI}} & \multicolumn{3}{c}{\textbf{ICEWS14}} & \multicolumn{3}{c}{\textbf{ICEWS18}} & \multicolumn{3}{c}{\textbf{ACLED-CD22}}\\
            \cmidrule(lr){2-4} \cmidrule(lr){5-7} \cmidrule(lr){8-10} \cmidrule(lr){11-13} \cmidrule(lr){14-16} & H@1 & H@3 & H@10 & H@1 & H@3 & H@10 & H@1 & H@3 & H@10 & H@1 & H@3 & H@10 & H@1 & H@3 & H@10 \\
            \midrule
\texttt{sLA-tKGF W/GPT-4 HL-5} &  0.801 &  0.838 &  0.866 &  0.750 &  0.799 &  0.827 &  0.475 &  0.584 &  0.753 &  0.515 &  0.617 &  0.717 &  0.450 &  0.584 &  0.685 \\
\texttt{sLA-tKGF W/GPT-4 HL-10} &  0.847 &  0.869 &  0.901 &  0.800 &  0.840 &  0.880 &  0.506 &  0.598 &  0.785 &  0.542 &  0.644 &  0.761 &  0.470 &  0.606 &  0.743 \\
\texttt{sLA-tKGF W/GPT-4 HL-15} &  0.885 &  0.926 &  0.946 &  0.845 &  0.877 &  0.930 &  0.530 &  0.648 &  0.823 &  0.572 &  0.683 &  0.788 &  0.502 &  0.639 &  0.780 \\
\texttt{sLA-tKGF W/GPT-4 HL-25} & \bf 0.961 & \bf 0.994 & \bf 1.000 & \bf 0.895 & \bf 0.943 & \bf 0.994 & \bf 0.566 & \bf 0.689 & \bf 0.879 & \bf 0.608 & \bf 0.726 & \bf 0.854 & \bf 0.537 & \bf 0.689 & \bf 0.822 \\
\texttt{sLA-tKGF W/GPT-4 HL-30} &  0.950 &  0.979 &  1.000 &  0.882 &  0.933 &  0.981 &  0.557 &  0.677 &  0.864 &  0.596 &  0.716 &  0.843 &  0.526 &  0.675 &  0.810 \\
            \bottomrule
        \end{tabular}
	}
	\end{minipage}
	\hfill
	    \begin{minipage}{1.0\textwidth}
	\centering
	\small
	\resizebox{1.0\textwidth}{!}{
		\begin{tabular}{ccccccccccccccccc}
            \toprule
            \multirow{1}{*}{\textbf{Multi-Step}} & \multicolumn{3}{c}{\textbf{YAGO}} & \multicolumn{3}{c}{\textbf{WIKI}} & \multicolumn{3}{c}{\textbf{ICEWS14}} & \multicolumn{3}{c}{\textbf{ICEWS18}} & \multicolumn{3}{c}{\textbf{ACLED-CD22}}\\
            \cmidrule(lr){2-4} \cmidrule(lr){5-7} \cmidrule(lr){8-10} \cmidrule(lr){11-13} \cmidrule(lr){14-16} & H@1 & H@3 & H@10 & H@1 & H@3 & H@10 & H@1 & H@3 & H@10 & H@1 & H@3 & H@10 & H@1 & H@3 & H@10 \\
            \midrule
\texttt{sLA-tKGF W/GPT-4 HL-5} &  0.765 &  0.804 &  0.842 &  0.728 &  0.774 &  0.803 &  0.423 &  0.527 &  0.695 &  0.454 &  0.579 &  0.698 &  0.414 &  0.553 &  0.643 \\
\texttt{sLA-tKGF W/GPT-4 HL-10} &  0.790 &  0.821 &  0.859 &  0.761 &  0.789 &  0.867 &  0.444 &  0.548 &  0.717 &  0.494 &  0.611 &  0.738 &  0.418 &  0.562 &  0.702 \\
\texttt{sLA-tKGF W/GPT-4 HL-15} &  0.842 &  0.875 &  0.895 &  0.812 &  0.833 &  0.899 &  0.459 &  0.591 &  0.747 &  0.518 &  0.630 &  0.761 &  0.448 &  0.600 &  0.743 \\
\texttt{sLA-tKGF W/GPT-4 HL-25} &  \bf 0.908 & \bf 0.938 & \bf 0.973 & \bf 0.846 & \bf 0.891 & \bf 0.968 & \bf 0.499 & \bf 0.633 & \bf 0.802 & \bf 0.545 & \bf 0.688 & \bf 0.828 & \bf 0.481 & \bf 0.648 & \bf 0.775 \\
\texttt{sLA-tKGF W/GPT-4 HL-30} &  0.900 &  0.922 &  0.962 &  0.833 &  0.881 &  0.952 &  0.489 &  0.624 &  0.791 &  0.531 &  0.676 &  0.817 &  0.470 &  0.635 &  0.761 \\
            \bottomrule
        \end{tabular}
	}
	\end{minipage}
    \vspace{0.1cm}
	\caption{The table illustrates the effect of using different historical context window sizes on tKG forecasting performance across several benchmark datasets. Increasing the historical context length consistently improved performance, indicating that more historical data enables the \texttt{sLA-tKGF} framework with GPT-4 integration to learn more effectively. The results highlighted in bold represent the performance of our proposed framework compared to existing methods.	
	}
	\label{tab:main-table8}
     \vspace{-0.15cm}
\end{table*}

\vspace{-2mm}
\begin{table*}[htbp]  
\centering
\resizebox{0.85\textwidth}{!}{
\begin{tabular}{c c c c c c c c c} \hline
Dataset&\#Training&\#Validation&\#Test&\#Entities& \#Relation& Time granularity &$N_{\text{obs}}$\\ \hline
ICEWS05-15$\_$continuous  & $148,673$ & $17,195$ & $17188$ & $10488$ & $251$ & NA & $1543$\\ 
\hline
\end{tabular}}
\vspace{2mm}
\caption{The table displays the statistics of the new dataset. \#Training, \#Validation, \#Test represent the number of quadruples in the training set, validation set, and test set, respectively. $N_{\text{obs}}$ represents the total number of snapshots in the new benchmark tKG forecasting dataset, where each snapshot captures the state of the tKG at a specific point in time.}
\label{tab:main-table10}
\vspace{-2mm}
\end{table*}

\vspace{-2mm}
\begin{table}[ht!]
\centering
    \begin{minipage}{0.70\textwidth}
	\small
	\resizebox{0.70\textwidth}{!}{
		\begin{tabular}{ccccccccccccccccc}
            \toprule
            \multirow{1}{*}{\textbf{Single-Step}} & \multicolumn{3}{c}{\textbf{ICEWS05-15$\_$continuous}} & \\
            \cmidrule(lr){2-4} \cmidrule(lr){5-7} \cmidrule(lr){8-10} \cmidrule(lr){11-13} \cmidrule(lr){14-16} & H@1 & H@3 & H@10 & \\
            \midrule
            \texttt{sLA-tKGF W/GPT-4} &  60.83 &  72.31 &  96.84  \\   
            \texttt{sLA-tKGF W/GPT-3.5} &  48.36 &  60.94 &  78.68  \\ 
            \texttt{sLA-tKGF text-davinci-003} &  43.43 &  53.86 &  71.64  \\   
            \texttt{sLA-tKGF W/Google Bard} &  54.06 &  68.53 &  90.90  \\
            \bottomrule
        \end{tabular}
	}
	\end{minipage}
	\hfill
	    \begin{minipage}{0.70\textwidth}
	\small
	\resizebox{0.70\textwidth}{!}{
		\begin{tabular}{ccccccccccccccccc}
            \toprule
            \multirow{1}{*}{\textbf{Multiple-Step}} & \multicolumn{3}{c}{\textbf{ICEWS05-15$\_$continuous}} & \\
            \cmidrule(lr){2-4} \cmidrule(lr){5-7} \cmidrule(lr){8-10} \cmidrule(lr){11-13} \cmidrule(lr){14-16} & H@1 & H@3 & H@10 & \\
            \midrule
            \texttt{sLA-tKGF W/GPT-4} &  51.32 &  65.17 &  87.23  \\   
            \texttt{sLA-tKGF W/GPT-3.5} &  42.15 &  49.84 &  65.59  \\ 
            \texttt{sLA-tKGF text-davinci-003} &  37.22 &  45.98 &  64.01  \\   
            \texttt{sLA-tKGF W/Google Bard} &  45.99 &  60.83 &  73.59  \\  
            \bottomrule
        \end{tabular}
	}
	\end{minipage}
    \vspace{0.1cm}
	\caption{The table shows the tKG forecasting results on Irregular temporal KGs.}
	\label{tab:main-table9}
     \vspace{-0.5cm}
\end{table}

\vspace{5mm}
\subsection{\textbf{Impact of Retrieved Fact Types}}
In this work, we introduce a zero-shot learning method to predict missing entities in query quadruples. Our approach includes: (a) Constructing a historical context for a query quadruple ($s_q, p_q, ?, t_q$) using historical facts from previous static KG snapshots $\mathcal{G}_{t_q-m:t_q-1}$ to form a knowledge-infused augmented prompt. (b) Converting retrieved facts and the query into verbalized sentences, employing sentence embedding for knowledge distillation, and using semantic similarity to filter facts, thereby constructing augmented prompts. (c) Estimating the missing entity conditioned on the augmented prompt, following $o_q \sim P(o \hspace{0.5mm} | \hspace{0.5mm}(s_q, r_q, ?, t_q), \mathcal{G}_{(t_q-m, t_q-1)})$. We evaluate the impact of \textit{single-subject entity} and \textit{subject entity-relation pair} historical facts on forecasting accuracy. The former involves only the subject entity $(s_q)$, while the latter includes both the subject $(s_q)$ and the relation $(r_q)$. We examine these effects across various benchmark KG datasets, analyzing the influence of different retrieved facts on the framework's performance. We find that \textit{single-subject entity} queries benefit from a broader range of historical facts, improving performance, while \textit{subject entity-relation pair} queries yield a more targeted set of facts, potentially enhancing outcomes. Our findings, as indicated in Table \ref{tab:main-table11}, demonstrate that the performance of the \texttt{sLA-tKGF W/GPT-4} framework varies depending on the dataset. The WIKI and ICEWS18 benchmarks exhibit improvements with entity-focused facts, whereas ICEWS14 performs better with pair-focused facts. Our study explores the impact of either $\textit{single-subject entity}$ or $\textit{subject-relation pair}$ facts, revealing that different datasets benefit from specific types of facts. This leads to more context-aware enhancements in the \texttt{sLA-tKGF W/GPT-4} framework.

\vspace{-1mm}
\begin{table*}[ht!]
  \vspace{-0.1cm}
    \begin{minipage}{1.0\textwidth}
	\centering
	\small
	\resizebox{1.0\textwidth}{!}{
		\begin{tabular}{ccccccccccccccccc}
            \toprule
            \multirow{1}{*}{\textbf{Single-Step}} & \multicolumn{3}{c}{\textbf{YAGO}} & \multicolumn{3}{c}{\textbf{WIKI}} & \multicolumn{3}{c}{\textbf{ICEWS14}} & \multicolumn{3}{c}{\textbf{ICEWS18}} & \multicolumn{3}{c}{\textbf{ACLED-CD22}}\\
            \cmidrule(lr){2-4} \cmidrule(lr){5-7} \cmidrule(lr){8-10} \cmidrule(lr){11-13} \cmidrule(lr){14-16} & H@1 & H@3 & H@10 & H@1 & H@3 & H@10 & H@1 & H@3 & H@10 & H@1 & H@3 & H@10 & H@1 & H@3 & H@10 \\
            \midrule
            \texttt{sLA-tKGF W/GPT-4}(uses both) &  \bf 0.961 & \bf 0.994 & \bf 1.000 & \bf 0.895 & \bf 0.943 & \bf 0.994 & \bf 0.566 & \bf 0.689 & \bf 0.879 & \bf 0.608 & \bf 0.726 & \bf 0.854 & \bf 0.537 & \bf 0.689 & \bf 0.822\\
            \texttt{sLA-tKGF W/GPT-4 (Single Entity)}  & 0.804 & 0.831 & 0.848 & 0.740 & 0.791 & 0.831 & 0.470 & 0.576 & 0.736 & 0.506 & 0.606 & 0.713 & 0.460 & 0.587 & 0.690  \\
            \texttt{sLA-tKGF W/GPT-4 (Entity Pair)} & 0.854 & 0.881 & 0.890 & 0.784 & 0.833 & 0.883 & 0.507 & 0.616 & 0.776 & 0.547 & 0.650 & 0.769 & 0.482 & 0.616 & 0.733  \\
            \bottomrule
        \end{tabular}
	}
	\end{minipage}
	\hfill
	    \begin{minipage}{1.0\textwidth}
	\centering
	\small
	 \resizebox{1.0\textwidth}{!}{
		\begin{tabular}{ccccccccccccccccc}
            \toprule
            \multirow{1}{*}{\textbf{Multi-Step}} & \multicolumn{3}{c}{\textbf{YAGO}} & \multicolumn{3}{c}{\textbf{WIKI}} & \multicolumn{3}{c}{\textbf{ICEWS14}} & \multicolumn{3}{c}{\textbf{ICEWS18}} & \multicolumn{3}{c}{\textbf{ACLED-CD22}}\\
            \cmidrule(lr){2-4} \cmidrule(lr){5-7} \cmidrule(lr){8-10} \cmidrule(lr){11-13} \cmidrule(lr){14-16} & H@1 & H@3 & H@10 & H@1 & H@3 & H@10 & H@1 & H@3 & H@10 & H@1 & H@3 & H@10 & H@1 & H@3 & H@10 \\
            \midrule
            \texttt{sLA-tKGF W/GPT-4} (uses both) &  \bf 0.908 & \bf 0.938 & \bf 0.973 & \bf 0.846 & \bf 0.891 & \bf 0.968 & \bf 0.499 & \bf 0.633 & \bf 0.802 & \bf 0.545 & \bf 0.688 & \bf 0.828 & \bf 0.481 & \bf 0.648 & \bf 0.775\\
            \texttt{sLA-tKGF W/GPT-4 (Single Entity)}  &  0.754 & 0.786 & 0.825 & 0.709 & 0.758 & 0.823 & 0.427 & 0.543 & 0.684 & 0.464 & 0.585 & 0.702 & 0.414 & 0.551 & 0.659  \\
            \texttt{sLA-tKGF W/GPT-4 (Entity Pair)} & 0.810 & 0.836 & 0.875 & 0.756 & 0.797 & 0.867 & 0.440 & 0.559 & 0.718 & 0.487 & 0.616 & 0.743 & 0.433 & 0.581 & 0.702  \\
            \bottomrule
        \end{tabular}
	}
	\end{minipage}
    \vspace{0.1cm}
	\caption{The table presents experimental results from a study on the impact of retrieved fact types on tKG forecasting performance. Two types of historical facts, `Single-Subject Entity' and `Subject Entity-Relation Pair', were evaluated for their impact on forecasting accuracy. Results indicated that the performance of the \texttt{sLA-tKGF W/GPT-4} framework varies based on the dataset and the type of historical fact used. In essence, the research highlights the importance of including both historical fact types in the augmented prompts for optimal performance in the forecasting task across various datasets.}
	\label{tab:main-table11}
    \vspace{-0.2cm}
\end{table*}

\vspace{-3mm}
\begin{table*}[ht!]
    \begin{minipage}{1.0\textwidth}
	\centering
	\small
	 \resizebox{1.0\textwidth}{!}{
		\begin{tabular}{ccccccccccccccccc}
            \toprule
            \multirow{1}{*}{\textbf{Single-Step}} & \multicolumn{3}{c}{\textbf{YAGO}} & \multicolumn{3}{c}{\textbf{WIKI}} & \multicolumn{3}{c}{\textbf{ICEWS14}} & \multicolumn{3}{c}{\textbf{ICEWS18}} & \multicolumn{3}{c}{\textbf{ACLED-CD22}}\\
            \cmidrule(lr){2-4} \cmidrule(lr){5-7} \cmidrule(lr){8-10} \cmidrule(lr){11-13} \cmidrule(lr){14-16} & H@1 & H@3 & H@10 & H@1 & H@3 & H@10 & H@1 & H@3 & H@10 & H@1 & H@3 & H@10 & H@1 & H@3 & H@10 \\
            \midrule
            \texttt{sLA-tKGF W/GPT-4 ($unidirectional$)}  &  \bf 0.961 & \bf 0.994 & \bf 1.000 & \bf 0.895 & \bf 0.943 & \bf 0.994 & \bf 0.566 & \bf 0.689 & \bf 0.879 & \bf 0.608 & \bf 0.726 & \bf 0.854 & \bf 0.537 & \bf 0.689 & \bf 0.822\\
            \texttt{sLA-tKGF W/GPT-4 ($bidirectional$)} &  0.971 & 1.000 & 1.00 & 0.920 & 0.969 & 1.000 & 0.585 & 0.712 & 0.905 & 0.628 & 0.754 & 0.881 & 0.538 & 0.707 & 0.844 \\
            \bottomrule
        \end{tabular}
	}
	\end{minipage}
	\hfill
	    \begin{minipage}{1.0\textwidth}
	\centering
	\small
	\resizebox{1.0\textwidth}{!}{
		\begin{tabular}{ccccccccccccccccc}
            \toprule
            \multirow{1}{*}{\textbf{Multi-Step}} & \multicolumn{3}{c}{\textbf{YAGO}} & \multicolumn{3}{c}{\textbf{WIKI}} & \multicolumn{3}{c}{\textbf{ICEWS14}} & \multicolumn{3}{c}{\textbf{ICEWS18}} & \multicolumn{3}{c}{\textbf{ACLED-CD22}}\\
            \cmidrule(lr){2-4} \cmidrule(lr){5-7} \cmidrule(lr){8-10} \cmidrule(lr){11-13} \cmidrule(lr){14-16} & H@1 & H@3 & H@10 & H@1 & H@3 & H@10 & H@1 & H@3 & H@10 & H@1 & H@3 & H@10 & H@1 & H@3 & H@10 \\
            \midrule
            \texttt{sLA-tKGF W/GPT-4 ($unidirectional$)}  &  \bf 0.908 & \bf 0.938 & \bf 0.973 & \bf 0.846 & \bf 0.891 & \bf 0.968 & \bf 0.499 & \bf 0.633 & \bf 0.802 & \bf 0.545 & \bf 0.688 & \bf 0.828 & \bf 0.481 & \bf 0.648 & \bf 0.775\\
            \texttt{sLA-tKGF W/GPT-4 ($bidirectional$)} &  0.930 & 0.965 & 0.999 & 0.875 & 0.915 & 0.982 & 0.502 & 0.650 & 0.824 & 0.564 & 0.707 & 0.852 & 0.496 & 0.665 & 0.797  \\
            \bottomrule
        \end{tabular}
	}
	\end{minipage}
    \vspace{0.1cm}
	\caption{The table presents the experimental results on the study of the impact of directionality in historical modeling on tKG forecasting performance. In the context of tKG forecasting, `unidirectional' and `bidirectional' indicate whether entities in historical facts align with the positions of the query quadruple. `Unidirectional' maintains alignment, while `bidirectional' allows entities to shift after transforming facts with inverse relations. The aim is to determine if incorporating both original historical facts and their inverse relations in the `bidirectional' modeling context can enhance the performance of tKG forecasting by offering a more varied historical context. The adoption of `bidirectional' relation modeling for tKG forecasting shows marginal improvements, with particularly notable gains evident in the ICEWS benchmark datasets.}
	\label{tab:main-table12}
    \vspace{-0.3cm}
\end{table*}

\subsection{\textbf{Impact of Directionality in Historical Modeling on tKG forecasting Performance}}
In the tKG forecasting task, `unidirectional' refers to when the subject entity ($s_q$) or subject-relation pair ($s_q, r_q$) from historical facts matches their position in the query quadruple $(s_q, r_q, ?, t_q)$. `Bidirectional' denotes cases where they can appear in any position. For bidirectional modeling, historical facts are transformed by swapping entities and inverting the relation, e.g., $(s, r, o, t)$ becomes $(o, r^{-1}, s, t)$. We use PLLMs such as GPT-4 to obtain reciprocal relations, enhancing tKG forecasting by incorporating diverse historical contexts. For instance, $\textit{(Barack Obama, visit, India, 2015-01-25)}$ becomes its reciprocal $\textit{(India, visited by, Barack Obama, 2015-01-25)}$. Our study evaluates the directionality's impact on the tKG forecasting, finding that bidirectional modeling slightly improves performance, notably on ICEWS datasets. The experimental results highlight the value of appropriate relation modeling for \texttt{sLA-tKGF W/GPT-4} in understanding context, offering modest performance boosts in tKG forecasting. Table \ref{tab:main-table12} shows the experimental results.

\vspace{-4mm}
\subsection{\textbf{Impact of PLLMs size}}
The \texttt{sLA-tKGF} framework leverages a blend of historical tKG data, current web-scraped information, and contextually relevant descriptions of past entity relationships generated by pre-trained language models (PLLMs) to construct knowledge-augmented prompts. These prompts query the small-scale language model to estimate forecasts. Designed to enhance both reliability and accountability, the framework offers a significant advancement over traditional forecasting methods. In this study, we examine the power law relationship between PLLM model size and performance on tKG forecasting using our \texttt{sLA-tKGF} framework. We explore the effect of PLLM model size on \texttt{sLA-tKGF} framework performance in zero-shot tKG forecasting through experiments with various PLLMs. Table \ref{tab:main-table13} lists the language models used, including GPT-2~\cite{radford2019language}, GPT-J~\cite{wang2021mesh}, and GPT-NeoX~\cite{black2022gpt}, all employing the GPT-2 BPE tokenizer~\cite{radford2019language} with similar vocabulary sizes. Our findings, detailed in Table \ref{tab:main-table15}, confirm that larger models yield better results, supporting the scaling laws in zero-shot learning. These models demonstrate enhanced linguistic comprehension, more complex architectures, and enhanced generalization capabilities.

\vspace{-5mm}
\begin{table}[ht!]
    \centering
    \resizebox{0.65\columnwidth}{!}{\begin{tabular}{cccc}
        \toprule
        Model Family & Model Name & \# Params  \\
        \midrule
        GPT2 & \texttt{gpt2} & 124M \\
             & \texttt{gpt2-xl} & 1.5B  \\
        \midrule
        GPT-J & \texttt{gpt-j-6b} & 6B  \\
        \midrule
        GPT-NeoX & \texttt{gpt-neox-20b} & 20B  \\
        \midrule
    \end{tabular}
    }
    \vspace{0mm}
    \caption{Overview of different GPT-based models by family and parameter count.}
    \label{tab:main-table13}
    \vspace{-8mm}
\end{table}

Our results highlight the advantages of using larger PLLMs resulting in \texttt{sLA-tKGF} framework with complex pattern recognition in tKG forecasting. With PLLMs like GPT-2, GPT-3, and GPT-4 showing remarkable abilities in natural language tasks, GPT-4 stands out for its scale and advanced learning capabilities. Yet, achieving success in tKG forecasting may require domain-specific expertise to develop effective zero-shot prompts.

\vspace{-3mm}
\begin{table*}[ht!]
    \begin{minipage}{1.0\textwidth}
	\centering
	\small
	\resizebox{1.0\textwidth}{!}{
		\begin{tabular}{ccccccccccccccccc}
            \toprule
            \multirow{1}{*}{\textbf{Single-Step}} & \multicolumn{3}{c}{\textbf{YAGO}} & \multicolumn{3}{c}{\textbf{WIKI}} & \multicolumn{3}{c}{\textbf{ICEWS14}} & \multicolumn{3}{c}{\textbf{ICEWS18}} & \multicolumn{3}{c}{\textbf{ACLED-CD22}}\\
            \cmidrule(lr){2-4} \cmidrule(lr){5-7} \cmidrule(lr){8-10} \cmidrule(lr){11-13} \cmidrule(lr){14-16} & H@1 & H@3 & H@10 & H@1 & H@3 & H@10 & H@1 & H@3 & H@10 & H@1 & H@3 & H@10 & H@1 & H@3 & H@10 \\
            \midrule
            \texttt{sLA-tKGF W/GPT-4} & \bf 0.961 & \bf 0.994 & \bf 1.000 & \bf 0.895 & \bf 0.943 & \bf 0.994 & \bf 0.566 & \bf 0.689 & \bf 0.879 & \bf 0.608 & \bf 0.726 & \bf 0.854 & \bf 0.537 & \bf 0.689 & \bf 0.822\\
            \texttt{sLA-tKGF W/GPT-2 ($\texttt{gpt2}$)}  &  0.505& 0.520& 0.513& 0.445& 0.487& 0.516& 0.243& 0.318& 0.437& 0.270& 0.345& 0.416& 0.223& 0.318& 0.398 \\
            \texttt{sLA-tKGF W/GPT-2 ($\texttt{gpt2-xl}$)} & 0.640& 0.661& 0.657& 0.569& 0.619& 0.656& 0.319& 0.413& 0.550& 0.349& 0.446& 0.541& 0.296& 0.407& 0.512 \\
            \texttt{sLA-tKGF W/GPT-J}  & 0.717& 0.734& 0.736& 0.643& 0.690& 0.734& 0.366& 0.466& 0.619& 0.396& 0.496& 0.602& 0.339& 0.469& 0.575 \\
            \texttt{sLA-tKGF GPT-NeoX} & 0.789& 0.804& 0.827& 0.719& 0.763& 0.811& 0.416& 0.525& 0.700& 0.452& 0.561& 0.681& 0.391& 0.532& 0.647 \\
            \bottomrule
        \end{tabular}
	}
	\end{minipage}
	\hfill
	    \begin{minipage}{1.0\textwidth}
	\centering
	\small
	\resizebox{1.0\textwidth}{!}{
		\begin{tabular}{ccccccccccccccccc}
            \toprule
            \multirow{1}{*}{\textbf{Multi-Step}} & \multicolumn{3}{c}{\textbf{YAGO}} & \multicolumn{3}{c}{\textbf{WIKI}} & \multicolumn{3}{c}{\textbf{ICEWS14}} & \multicolumn{3}{c}{\textbf{ICEWS18}} & \multicolumn{3}{c}{\textbf{ACLED-CD22}}\\
            \cmidrule(lr){2-4} \cmidrule(lr){5-7} \cmidrule(lr){8-10} \cmidrule(lr){11-13} \cmidrule(lr){14-16} & H@1 & H@3 & H@10 & H@1 & H@3 & H@10 & H@1 & H@3 & H@10 & H@1 & H@3 & H@10 & H@1 & H@3 & H@10 \\
            \midrule
            \texttt{sLA-tKGF W/GPT-4} & \bf 0.908 & \bf 0.938 & \bf 0.973 & \bf 0.846 & \bf 0.891 & \bf 0.968 & \bf 0.499 & \bf 0.633 & \bf 0.802 & \bf 0.545 & \bf 0.688 & \bf 0.828 & \bf 0.481 & \bf 0.648 & \bf 0.775\\
            \texttt{sLA-tKGF W/GPT-2 ($\texttt{gpt2}$)}  & 0.479 & 0.485 & 0.509 & 0.438 & 0.469 & 0.485 & 0.228 & 0.304 & 0.425 & 0.267 & 0.335 & 0.416 & 0.220 & 0.301 & 0.395 \\
            \texttt{sLA-tKGF W/GPT-2 ($\texttt{gpt2-xl}$)} &  0.539 & 0.558 & 0.576 & 0.498 & 0.524 & 0.558 & 0.273 & 0.355 & 0.481 & 0.299 & 0.378 & 0.465 & 0.253 & 0.353 & 0.442 \\
            \texttt{sLA-tKGF W/GPT-J}  & 0.603 & 0.637 & 0.648 & 0.564 & 0.597 & 0.641 & 0.315 & 0.411 & 0.554 & 0.349 & 0.443 & 0.521 & 0.296 & 0.410 & 0.497 \\
            \texttt{sLA-tKGF GPT-NeoX} & 0.691 & 0.716 & 0.732 & 0.635 & 0.670 & 0.709 & 0.361 & 0.463 & 0.617 & 0.395 & 0.498 & 0.602 & 0.338 & 0.464 & 0.576 \\
            \bottomrule
        \end{tabular}
	}
	\end{minipage}
    \vspace{0.cm}
	\caption{The table presents the experimental results of the study on the impact of PLLM size on the tKG forecasting task.}
	\label{tab:main-table15}
    \vspace{-0.2cm}
\end{table*}

\vspace{-0mm}
\begin{table*}[ht!]
    \begin{minipage}{1.0\textwidth}
	\centering
	\small
	\resizebox{1.0\textwidth}{!}{
		\begin{tabular}{ccccccccccccccccc}
            \toprule
            \multirow{1}{*}{\textbf{Single-Step}} & \multicolumn{3}{c}{\textbf{YAGO}} & \multicolumn{3}{c}{\textbf{WIKI}} & \multicolumn{3}{c}{\textbf{ICEWS14}} & \multicolumn{3}{c}{\textbf{ICEWS18}} & \multicolumn{3}{c}{\textbf{ACLED-CD22}}\\
            \cmidrule(lr){2-4} \cmidrule(lr){5-7} \cmidrule(lr){8-10} \cmidrule(lr){11-13} \cmidrule(lr){14-16} & H@1 & H@3 & H@10 & H@1 & H@3 & H@10 & H@1 & H@3 & H@10 & H@1 & H@3 & H@10 & H@1 & H@3 & H@10 \\
            \midrule
            \texttt{$\qquad (b = 48, e_{P} = 30, d = 64)$} & 0.903 & 0.940 & 0.953 & 0.844 & 0.868 & 0.907 &  0.492 &  0.620 & 0.802 & 0.541 & 0.655 & 0.792 & 0.487 & 0.641 &  0.757\\
\texttt{$\qquad (b = 48, e_{P} = 30, d = 128)$}  & \bf 0.961 & \bf 0.994 & \bf 1.000 & \bf 0.895 & \bf 0.943 & \bf 0.994 & \bf 0.566 & \bf 0.689 & \bf 0.879 & \bf 0.608 & \bf 0.726 & \bf 0.854 & \bf 0.537 & \bf 0.689 & \bf 0.822\\
\texttt{$\qquad (b = 16, e_{P} = 30, d = 64)$} & 0.881 & 0.934 & 0.938 & 0.820 & 0.853 & 0.895 &  0.465 &  0.599 & 0.787 & 0.514 & 0.644 & 0.781 & 0.472 & 0.596 &  0.740\\         
            \bottomrule
        \end{tabular}
	}
	\end{minipage}
	\hfill
	    \begin{minipage}{1.0\textwidth}
	\centering
	\small
	\resizebox{1.0\textwidth}{!}{
		\begin{tabular}{ccccccccccccccccc}
            \toprule
            \multirow{1}{*}{\textbf{Multi-Step}} & \multicolumn{3}{c}{\textbf{YAGO}} & \multicolumn{3}{c}{\textbf{WIKI}} & \multicolumn{3}{c}{\textbf{ICEWS14}} & \multicolumn{3}{c}{\textbf{ICEWS18}} & \multicolumn{3}{c}{\textbf{ACLED-CD22}}\\
            \cmidrule(lr){2-4} \cmidrule(lr){5-7} \cmidrule(lr){8-10} \cmidrule(lr){11-13} \cmidrule(lr){14-16} & H@1 & H@3 & H@10 & H@1 & H@3 & H@10 & H@1 & H@3 & H@10 & H@1 & H@3 & H@10 & H@1 & H@3 & H@10 \\
            \midrule
            \texttt{$\qquad (b = 48, e_{P} = 30, d = 64)$} & 0.849 & 0.902 & 0.920 & 0.808 & 0.822 & 0.934 & 0.465 & 0.603 & 0.750 & 0.490 & 0.651 & 0.790 & 0.469 & 0.612 &  0.718\\
            \texttt{$\qquad (b = 48, e_{P} = 30, d = 128)$}  & \bf 0.908 & \bf 0.938 & \bf 0.973 & \bf 0.846 & \bf 0.891 & \bf 0.968 & \bf 0.499 & \bf 0.633 & \bf 0.802 & \bf 0.545 &                    \bf 0.688 & \bf 0.828 & \bf 0.481 & \bf 0.648 & \bf 0.775\\
             \texttt{$\qquad (b = 16, e_{P} = 30, d = 64)$} & 0.842 & 0.891 & 0.903 & 0.799 & 0.836 & 0.909 & 0.452 & 0.584 & 0.755 & 0.504 & 0.637 & 0.795 & 0.465 & 0.902 &  0.703\\
            \bottomrule
        \end{tabular}
	}
	\end{minipage}
    \vspace{0.1cm}
	\caption{The table presents the hyperparameter tuning results conducted on the tKG forecasting task using benchmark datasets.}
	\label{tab:main-table15}
    \vspace{-0.2cm}
\end{table*} 

\vspace{-2mm}
\begin{table*}[ht!]
    \begin{minipage}{1.0\textwidth}
	\centering
	\small
	 \resizebox{1.0\textwidth}{!}{
		\begin{tabular}{ccccccccccccccccc}
            \toprule
            \multirow{1}{*}{\textbf{Single-Step}} & \multicolumn{3}{c}{\textbf{YAGO}} & \multicolumn{3}{c}{\textbf{WIKI}} & \multicolumn{3}{c}{\textbf{ICEWS14}} & \multicolumn{3}{c}{\textbf{ICEWS18}} & \multicolumn{3}{c}{\textbf{ACLED-CD22}}\\
            \cmidrule(lr){2-4} \cmidrule(lr){5-7} \cmidrule(lr){8-10} \cmidrule(lr){11-13} \cmidrule(lr){14-16} & H@1 & H@3 & H@10 & H@1 & H@3 & H@10 & H@1 & H@3 & H@10 & H@1 & H@3 & H@10 & H@1 & H@3 & H@10 \\
            \midrule
\texttt{$(Top-P = 1, Temp = 0)$} & \bf 0.961 & \bf 0.994 & \bf 1.000 & \bf 0.895 & \bf 0.943 & \bf 0.994 & \bf 0.566 & \bf 0.689 & \bf 0.879 & \bf 0.608 & \bf 0.726 & \bf 0.854 & \bf 0.537 & \bf 0.689 & \bf 0.822\\
\texttt{$(Top-P = 1, Temp = 3)$} & 0.832 & 0.879 & 0.908 & 0.776 & 0.806 & 0.896 & 0.489 & 0.616 & 0.776 & 0.534 & 0.624 & 0.733 & 0.491 & 0.602 & 0.729\\
\texttt{$(Top-P = 0, Temp = 0)$} & 0.909 & 0.932 & 0.954 & 0.813 & 0.881 & 0.959 & 0.529 & 0.660 & 0.822 & 0.569 & 0.688 & 0.817 & 0.495 & 0.670 & 0.761\\
            \bottomrule
        \end{tabular}
	}
	\end{minipage}
	\hfill
	    \begin{minipage}{1.0\textwidth}
	\centering
	\small
	\resizebox{1.0\textwidth}{!}{
		\begin{tabular}{ccccccccccccccccc}
            \toprule
            \multirow{1}{*}{\textbf{Multi-Step}} & \multicolumn{3}{c}{\textbf{YAGO}} & \multicolumn{3}{c}{\textbf{WIKI}} & \multicolumn{3}{c}{\textbf{ICEWS14}} & \multicolumn{3}{c}{\textbf{ICEWS18}} & \multicolumn{3}{c}{\textbf{ACLED-CD22}}\\
            \cmidrule(lr){2-4} \cmidrule(lr){5-7} \cmidrule(lr){8-10} \cmidrule(lr){11-13} \cmidrule(lr){14-16} & H@1 & H@3 & H@10 & H@1 & H@3 & H@10 & H@1 & H@3 & H@10 & H@1 & H@3 & H@10 & H@1 & H@3 & H@10 \\
            \midrule
\texttt{$(Top-P = 1, Temp = 0)$} & \bf 0.908 & \bf 0.938 & \bf 0.973 & \bf 0.846 & \bf 0.891 & \bf 0.968 & \bf 0.499 & \bf 0.633 & \bf 0.802 & \bf 0.545 & \bf 0.688 & \bf 0.828 & \bf 0.481 & \bf 0.648 & \bf 0.775\\
\texttt{$(Top-P = 1, Temp = 3)$} & 0.813 & 0.828 & 0.832 & 0.742 & 0.799 & 0.819 & 0.442 & 0.564 & 0.688 & 0.478 & 0.580 & 0.750 & 0.412 & 0.576 & 0.686\\
\texttt{$(Top-P = 0, Temp = 0)$} & 0.859 & 0.885 & 0.908 & 0.786 & 0.833 & 0.927 & 0.470 & 0.607 & 0.749 & 0.503 & 0.659 & 0.775 & 0.450 & 0.613 & 0.727\\   
            \bottomrule
        \end{tabular}
	}
	\end{minipage}
    \vspace{0.1cm}
	\caption{The table shows the experimental results of hyperparameter optimization for both Top-P and temperature conducted on the tKG forecasting task using benchmark datasets, specifically for the \texttt{sLA-tKGF-GPT-4} variant of our framework.}
	\label{tab:main-table16}
     \vspace{-0.5cm}
\end{table*} 

\vspace{0.5mm}
\subsection{\textbf{Hyperparameter Tuning}}
Our \texttt{sLA-tKGF-GPT-4} framework improves tKG forecasting accuracy by integrating PLLMs with small-scale language models. Hyperparameter tuning is challenging due to its complex dimensionality, computational demands, and dataset-specific requirements. We opt for random search over grid search or Bayesian optimization for efficient hyperparameter exploration, seeking the optimal configuration on benchmark datasets. The small-scale language models are trained on knowledge-augmented prompting for tKG forecasting, aiming to predict missing entities by minimizing cross-entropy loss. Hyperparameter optimization for \texttt{sLA-tKGF-GPT-4} focuses on batch size ($b \in {16, 48}$), epochs ($e_p \in {10, 30}$), and embedding dimension ($d \in {64, 128}$). Table \ref{tab:main-table15} shows tuning results on benchmark datasets, highlighting that the combination ($b = 48, e_p = 30, d = 128$) is most effective for tKG forecasting tasks.

\vspace{-0.3cm}
\subsection{\textbf{Hyperparameter Tuning of PLLMs}}
Top-P and temperature serve as hyperparameters in pre-trained large language models (PLLMs), such as GPT-4 and Google Gemini. These hyperparameters influence the predictability and variety of model outputs. The temperature parameter primarily affects output predictability. Lower temperature values result in more deterministic responses, while higher values enable greater variability. A temperature value of 1.0 maintains the original output probabilities. On the other hand, Top-P (nucleus) sampling dynamically chooses tokens based on their cumulative probability. This method strikes a balance between text generation diversity and coherence by adjusting the threshold. By fine-tuning these parameters, it is possible to strike an optimal balance between randomness and determinism in generated outputs. Regarding the \texttt{sLA-tKGF-GPT-4} variant, hyperparameter optimization studies indicate that Top-P ranges over $[0, 1]$, and temperature falls within the range of $[0, 3]$. As shown in Table \ref{tab:main-table16}, top performance has been consistently observed when using the configuration $(\text{Top-P}=1,\text{Temp}=0)$ across multiple benchmark datasets, highlighting its broad applicability and efficacy.

\end{document}